\documentclass[pmlr,twocolumn]{jmlr}

\usepackage[utf8]{inputenc}
\usepackage{color}
\usepackage{amsfonts}
\usepackage{bm}
\usepackage{enumitem}
\usepackage{multirow}
\usepackage{multicol}

\usepackage{dsfont}
\usepackage{cancel}

\usepackage{threeparttable}
\usepackage{wrapfig}
\usepackage{multirow}
\usepackage{tcolorbox}
\usepackage{makecell}
\usepackage[section]{algorithm}%
\SetKwInOut{Parameter}{parameter}
\usepackage{color}
\usepackage{xcolor}
\usepackage{float}

\def\pAp@k{\texttt{pAp@k}}

\def\p{\mathbf{p}}
\def\v{\mathbf{v}}
\def\u{\mathbf{u}}

\usepackage{colortbl}
\usepackage{collcell,xcolor,xfp}
\def\1{\cellcolor{green!30}}
\usepackage{tikz}
\newcommand\mybox[2][]{\tikz[overlay]\node[fill=blue!20,inner sep=2pt, anchor=text, rectangle, rounded corners=1mm,#1] {#2};\phantom{#2}}

\newcommand{\alphamin}{\alpha}

\newcommand{\fmtnum}[1]{%
  \ifnum\fpeval{#1 < 0} = 1
    \textcolor{red}{$#1$}%
  \else
    \ifnum\fpeval{#1 < 0.5} = 1
      \textcolor{green}{$#1$}%
    \else
      \textcolor{blue}{$#1$}%
    \fi
  \fi
}

\newcommand{\tableCox}{Cox}
\newcommand{\tableCoxKeyaInd}{Cox$_I$(Keya~et~al.)}
\newcommand{\tableCoxRPInd}{Cox$_I$(R\&P)}
\newcommand{\tableCoxKeyaGroup}{Cox$_G$(Keya~et~al.)}
\newcommand{\tableCoxRPGroup}{Cox$_G$(R\&P)}
\newcommand{\tableCoxKeyaInt}{Cox$_{\cap}$(Keya~et~al.)}

\newcommand{\tableDeepSurvKeyaInd}{DeepSurv$_I$(Keya~et~al.)}
\newcommand{\tableDeepSurvRPInd}{DeepSurv$_I$(R\&P)}
\newcommand{\tableDeepSurvKeyaGroup}{DeepSurv$_G$(Keya~et~al.)}
\newcommand{\tableDeepSurvRPGroup}{DeepSurv$_G$(R\&P)}
\newcommand{\tableDeepSurvKeyaInt}{DeepSurv$_{\cap}$(Keya~et~al.)}

\newcommand{\tableDROCox}{\textsc{DRO-COX}}
\newcommand{\tableDROCoxSplit}{\textsc{DRO-COX (SPLIT)}}
\newcommand{\tableDeepDROCox}{Deep \textsc{DRO-COX}}
\newcommand{\tableDeepDROCoxSplit}{Deep \textsc{DRO-COX (SPLIT)}}

\usepackage{longtable}%

\usepackage{booktabs}
\usepackage[load-configurations=version-1]{siunitx} %

\theorembodyfont{\upshape}
\theoremheaderfont{\scshape}
\theorempostheader{:}
\theoremsep{\newline}

\jmlrpages{}
\jmlrvolume{}

\jmlryear{2022}
\jmlrworkshop{Machine Learning for Health (ML4H) 2022}

\title[Distributionally Robust Survival Analysis]{Distributionally Robust Survival Analysis:\titlebreak A Novel Fairness Loss Without Demographics}

\makeatletter
\newcommand{\printfnsymbol}[1]{%
  \textsuperscript{\@fnsymbol{#1}}%
}
\makeatother

  \author{\Name{Shu Hu\nametag{\thanks{equal contribution}}} \Email{shuhu@cmu.edu}\\
  \Name{George H. Chen\nametag{$^*$}\nametag{\thanks{corresponding author}}} \Email{georgechen@cmu.edu}\\
  \addr Heinz College of Information Systems and Public Policy, Carnegie Mellon University\vspace{-.5em}}

\begin{document}

\maketitle

\begin{abstract}
We propose a general approach for training survival analysis models that minimizes a worst-case error across \emph{all} subpopulations that are large enough (occurring with at least a user-specified minimum probability). This approach uses a training loss function that does not know any demographic information to treat as sensitive. Despite this, we demonstrate that our proposed approach often scores better on recently established fairness metrics (without a significant drop in prediction accuracy) compared to various baselines, including ones which directly use sensitive demographic information in their training loss. Our code is available at: \url{https://github.com/discovershu/DRO_COX}
\end{abstract}

\begingroup
\setlength{\abovedisplayskip}{3pt plus 2pt}
\setlength{\belowdisplayskip}{3pt plus 2pt}

\vspace{-.75em}
\section{Introduction}
\label{sec:intro}
\vspace{-.25em}

One of the recent advances for encouraging fairness in machine learning models is to minimize a worst-case error over \emph{all} subpopulations that are large enough (e.g., \citealt{hashimoto2018fairness,duchi2021learning,li2021evaluating,duchi2022distributionally,hu2022rank}). In particular, a modeler specifies a probability threshold $\alpha$ of a minority subpopulation occurring. The goal is to ensure that all subpopulations with at least occurrence probability $\alpha$ have low error whereas we make no promises for subpopulations occurring with probability less than $\alpha$. The modeler need not provide a list of subpopulations to account for. This problem is tractable to solve in practice and is called \emph{distributionally robust optimization} (DRO).

We emphasize that curating a list of all subpopulations to account for can be challenging in practice for numerous reasons. For example, one major challenge is \emph{intersectionality}: subpopulations that a machine learning model yields the worst accuracy scores for can be defined by complex intersections of sensitive attributes (e.g., age, race, gender) \citep{buolamwini2018gender}. Some of these attributes might require discretization (e.g., dividing age into bins), for which choosing the ``best'' discretization strategy might not be straightforward. Moreover, if there is a large number of features and we suspect that the sensitive attributes (encoded by specific features) could possibly be correlated with other features (not flagged as sensitive), there is a question of whether these other features should also be accounted for in a listing of what the sensitive attributes are. DRO provides a theoretically sound alternative to having to specify such sensitive attributes in a training loss function.

Our main contribution in this paper is to show how to apply DRO to survival analysis. The key technical challenge is that existing DRO theory assumes that the overall training loss can be separated across individuals so that any individual's loss term does not depend on other individuals. This assumption does not hold for many survival analysis loss functions, including that of the popular Cox proportional hazards model \citep{cox1972regression}, due to pairwise comparisons from ranking or similarity score evaluations (e.g., \citealt{steck2007ranking,lee2018deephit,chen2020deep,wu2021uncertainty}). We use a sample splitting approach to address this technical challenge. We specifically show how to use DRO with the Cox model and its deep neural network variant \citep{faraggi1995neural, katzman2018deepsurv}. On three standard survival analysis datasets that have been previously used for research on fairness, our approach often outperforms various baseline methods in terms of existing fairness metrics that focus on user-specified sensitive attributes, including baselines with training loss functions that directly use these sensitive attributes (whereas ours does not). As with other fairness methods recently developed for survival analysis (e.g., \citet{keya2021equitable,rahman2022fair}), our approach also results in a drop in accuracy (compared to using a loss that does not encourage fairness). This tradeoff in accuracy vs fairness can be tuned by the user. For ease of presentation, we apply DRO only to classical and deep Cox models, but the ideas we use readily extend to other survival models as well.

\vspace{-.5em}
\section{Background}

We review the standard survival analysis setup in Section~\ref{sec:survival-analysis}, classical and neural network variants of the Cox proportional hazards model in Section~\ref{sec:cox-models}, and existing work on fair survival analysis in Section~\ref{sec:related_work}. We defer explaining the basics of DRO to Section~\ref{sec:method} when we simultaneously explain how we apply DRO to survival analysis.

\vspace{-.5em}
\subsection{Survival Analysis Setup}
\label{sec:survival-analysis}
\vspace{-.2em}

Survival analysis aims to model the amount of time that will elapse before a critical event of interest happens. Classically, this critical event is death (i.e., we model time until different individuals are deceased), but the critical event need not be death and could instead be, for example, discharge from a hospital, or awakening from a coma.

We assume that we have training data $\{(X_i,Y_i,\delta_i)\}_{i=1}^n$, where the $i$-th training patient has feature vector $X_i\in\mathcal{X}$, observed duration $Y_i\ge0$, and event indicator $\delta_i\in\{0,1\}$. If $\delta_i=1$ (i.e., the critical event of interest happened for the $i$-th patient), then $Y_i$ is the time until the event happens. Otherwise, if $\delta_i=0$, then $Y_i$ is the time until censoring for the $i$-th patient, i.e., the true time until event is unknown but we know that it is at least $Y_i$. In more detail, each training data point $(X_i,Y_i,\delta_i)$ is assumed to be generated from the following procedure:
\begin{enumerate}[itemsep=0pt,topsep=2pt,parsep=0pt,leftmargin=*,partopsep=0em] %
\item Sample feature vector $X_i$ from a feature vector distribution $\mathbb{P}_{X}$.
\item Sample nonnegative time duration $T_i$ (this is the true time until the critical event happens) from a conditional distribution $\mathbb{P}_{T|X=X_i}$.
\item Sample nonnegative time duration $C_i$ (this is the true time until the data point is censored) from a conditional distribution $\mathbb{P}_{C|X=X_i}$.
\item If $T_i \le C_i$ (the critical event happens before censoring), then set $Y_i = T_i$ and $\delta_i=1$. Otherwise, set $Y_i = C_i$ and $\delta_i=0$.
\end{enumerate}
Distributions $\mathbb{P}_X$, $\mathbb{P}_{T|X}$, and $\mathbb{P}_{C|X}$ are shared across data points and are unknown. %
We assume that the random variables $T_i$ and $C_i$ are independent given $X_i$. %
We denote the CDF of distribution $\mathbb{P}_{T|X=x}$ as $F(\cdot|x)$. %

A standard prediction task is to estimate the probability that a patient with feature vector~$x$ survives beyond time $t$. Formally, this is defined as the survival function
\begin{equation*}
S(t|x) := \mathbb{P}(T > t | X=x) = 1 - F(t|x). %
\end{equation*}
We explain how to estimate $S(\cdot|x)$ using variants of the Cox model next.

\subsection{Classical and Deep Cox Models}
\label{sec:cox-models}

The Cox proportional hazards model \citep{cox1972regression} estimates a transformed version of the survival function $S(\cdot|x)$ called the \emph{hazard function}, given by $h(t|x):= - \frac{\partial}{\partial t}\log S(t|x)$; from negating both sides of this equation, integrating over time, and exponentiating, we get $S(t|x)=\exp(-\int_0^t h(u|x)du)$. Thus, if we have an estimate of $h(\cdot|x)$, then we can readily estimate the survival function $S(\cdot|x)$.

The Cox model assumes that the hazard function has the factorization
\begin{equation}
h(t|x) = h_0(t)\exp(f(x;\theta)),
\label{eq:hazard-factorization}
\end{equation}
where $h_0$ is called the baseline hazard function ($h_0$ maps a nonnegative time $t\ge0$ to a nonnegative number), and $f(\cdot;\theta)$ is the so-called log partial hazard function ($f(x;\theta)$ could be thought of as assigning a real-valued ``risk score'' to feature vector~$x$: when $f(x;\theta)$ is higher, then~$x$ has a higher risk of the critical event happening); note that $\theta$ refers to the parameters of $f$.

\endgroup
\begingroup
\setlength{\abovedisplayskip}{2pt plus 0.5pt}
\setlength{\belowdisplayskip}{2pt plus 0.5pt}

The original Cox model \citep{cox1972regression} defines $f$ to be a dot product: $f(x;\theta)=\theta^T x$, where $\theta$ and $x$ are in the same Euclidean vector space. More recently, researchers replaced $f$ with a neural network \citep{faraggi1995neural,katzman2018deepsurv}, resulting in a method called DeepSurv. In either case, the standard approach for learning a Cox model is to first learn $f(\cdot;\theta)$ (i.e., learn the parameters~$\theta$) by minimizing the negative log partial likelihood:
\begin{equation}
\mathcal{L}_{\text{average}}(\theta) = \frac{1}{n} \sum_{i=1}^n \ell_i(\theta),
\label{eq:average}
\end{equation}
where the $i$-th patient's loss is
\begin{align}
&\ell_i(\theta) \nonumber \\
&:=-\delta_i \Big[f(X_i;\theta)-\log\!\!\sum_{\substack{j=1,\dots,n\\\text{s.t.~}Y_j\geq Y_i}}\!\!\exp(f(X_j;\theta))\Big]. \nonumber \\[-1.2em] \label{eq:cox-individual-loss}
\end{align}
If the $i$-th patient is censored ($\delta_i=0$), then $\ell_i(\theta)=0$. Thus, the loss $\mathcal{L}_{\text{average}}(\theta)$ weights \emph{uncensored} training patients equally. After learning $f(\cdot;\theta)$, we then estimate~$h_0$; as this step is not essential to our exposition, we explain it in Appendix~\ref{sec:breslow}, along with details on constructing the final estimate of $S(\cdot|x)$.

\vspace{-.25em}
\vspace{-.25em}
\vspace{-.5em}
\subsection{Fair Survival Analysis}\label{sec:related_work}
\vspace{-.25em}

Despite many advances in survival analysis methodology in recent years (e.g., see the survey by \citet{wang2019machine}), very few of these advances focus on fairness \citep{keya2021equitable, zhang2022longitudinal, sonabend2022flexible, rahman2022fair}. From a practical standpoint, asking for a survival analysis model to be fair is not different from asking for any other machine learning model to be fair in that if the model is used to assist high-stakes decision making (e.g., helping clinicians decide on personalized treatments, improving how hospitals allocate resources for different patients), then accounting for some notion of fairness could be an important design consideration. %

To this end, \citet{keya2021equitable} adapted existing fairness definitions to the survival analysis setting and showed how to encourage different notions of fairness by adding fairness regularization terms to the conventional loss function stated in equation~\eqref{eq:average}. Specifically, \citet{keya2021equitable} came up with individual \citep{dwork2012fairness}, group \citep{dwork2012fairness}, and intersectional \citep{foulds2020intersectional} fairness definitions specialized to Cox models. Keya et al.~define individual fairness in terms of model predictions being similar for similar individuals, and group fairness in terms of different user-specified groups having similar average predicted outcomes. Intersectional fairness further considers subgroups defined by intersections of protected groups (e.g., individuals of a specific race and simultaneously a specific gender) with the idea that intersections of protected groups could be vulnerable to additional harms.

However, a major limitation of the notions of fairness defined by \citet{keya2021equitable} for survival analysis is that they focus on predicted model outputs and do not actually use any of the label information (the observed time $Y_i$ and event indicator $\delta_i$ variables). For example, if one uses age as a sensitive attribute and suppose we discretize age into two groups, then the notion of group fairness by \citet{keya2021equitable} would be asking for the predicted outcomes of the two different age groups to be similar, which for healthcare problems often does not make sense (since age is often highly predictive of different health outcomes). Instead, in such a scenario, a more desirable notion of fairness is that the model's \emph{accuracy} for the different age groups be similar.

To account for model accuracy, \citet{zhang2022longitudinal} introduced a fairness metric called \emph{concordance imparity} that computes a quantity similar to the standard survival analysis accuracy metric of \emph{concordance index} \citep{harrell1982evaluating} for different groups and then looks at the worst-case difference between any two groups' accuracy scores. Meanwhile, \citet{rahman2022fair} directly modified the fairness definitions of \citet{keya2021equitable} to account for observed times and censoring information, and also generalized these definitions to survival models beyond Cox models.

Separately, \citet{sonabend2022flexible} empirically explored how well existing survival analysis accuracy and calibration metrics measure bias by synthetically modifying datasets (e.g., undersampling disadvantaged groups). However, they do not propose any new fairness metric or survival model that encourages fairness.

The papers mentioned above that propose new methods for learning fair survival models all either require user-specified demographic information to treat as sensitive (possibly as a list of subpopulations/groups to account for) or are simply adding a loss term that encourages smoothness in the model outputs (the individual fairness metrics by \citet{keya2021equitable} and \citet{rahman2022fair} are simply encouraging the predicted model output to be Lipschitz continuous; for details, see Appendix~\ref{sec:fairness-measures}). In contrast, our proposed approach does not require the user to specify any sensitive demographic attributes in the training loss function, and is not simply encouraging the model output to be Lipschitz continuous.

\endgroup
\begingroup
\setlength{\abovedisplayskip}{3pt plus 1pt}
\setlength{\belowdisplayskip}{3pt plus 1pt}

\vspace{-.75em}
\section{DRO for Survival Analysis}
\label{sec:method}
\vspace{-.3em}

We now present our proposed method that applies distributionally robust optimization (DRO) to survival analysis. DRO uses a worst-case average error over ``large enough'' subpopulations. %
Note that there are now a number of DRO variants (e.g., \citealt{hashimoto2018fairness,sagawa2020distributionally,duchi2021learning,duchi2022distributionally}). We use the one by \citet{hashimoto2018fairness}.

\smallskip
\noindent
\textbf{Problem setup.}
Let $\mathbb{P}$ denote the joint distribution over each data point $(X_i,Y_i,\delta_i)$. This joint distribution corresponds to the generative procedure described in Section~\ref{sec:survival-analysis}. We assume that there are $K$ groups that comprise $\mathbb{P}$. In particular, $\mathbb{P}$ is a mixture of $K$ distributions $\mathbb{P}:=\sum_{k=1}^{K}\pi_{k}\mathbb{P}_{k}$, where the $k$-th group occurs with probability $\pi_{k}\in(0,1)$ and has associated distribution $\mathbb{P}_{k}$. Moreover, $\sum_{k=1}^{K}\pi_{k}=1$. We assume that we do not know $\{(\pi_{k},\mathbb{P}_{k})\}_{k=1}^{K}$, nor do we know $K$. This setting, for instance, handles the case where we do not exhaustively know all subpopulations to consider. The smallest minority group corresponds to whichever group has the smallest $\pi_k$ value.

We would like to minimize the risk
\[
\mathcal{R}_{\max}(\theta):=\max_{k=1,\dots,K}\mathbb{E}_{(X,Y,\delta)\sim\mathbb{P}_{k}}[\widetilde{\ell}(\theta;X,Y,\delta)],
\]
where $\widetilde{\ell}$ is a loss function that depends only on the parameters $\theta$ (for a survival analysis model that we aim to learn) and on a single data point $(X,Y,\delta)$. However, minimizing $\mathcal{R}_{\max}(\theta)$ is not possible as we do not know any of the latent groups nor how many such groups there are. However, it turns out that there is an optimization problem that we can tractably solve that minimizes an empirical version of an upper bound on $\mathcal{R}_{\max}(\theta)$. We explain what the upper bound is in Section~\ref{sec:dro-risk-upper-bound}, how to empirically minimize the upper bound in Section~\ref{sec:empirical-dro-risk-minimization}, and finally how to choose the loss $\widetilde{\ell}$ in Section~\ref{sec:individual-loss}. Note that the material in Sections~\ref{sec:dro-risk-upper-bound} and~\ref{sec:empirical-dro-risk-minimization} is not novel; these sections translate the DRO formulation by \citet{hashimoto2018fairness} to survival analysis. On the other hand, Section~\ref{sec:individual-loss} is novel and focuses on a technical complication in applying DRO to survival analysis.

\vspace{-.5em}
\subsection{Upper Bound on the Risk \texorpdfstring{$\mathcal{R}_{\max}(\theta)$}{R\_\{max\}(theta)} Using DRO}
\label{sec:dro-risk-upper-bound}
\vspace{-.25em}

For a set of distributions $\mathcal{B}_{r}(\mathbb{P})$ to be defined shortly, we consider minimizing the following alternative risk instead:
\begin{equation}
\mathcal{R}_{\text{DRO}}(\theta;r):=\!\!\sup_{\mathbb{Q}\in\mathcal{B}_{r}(\mathbb{P})}\!\!\mathbb{E}_{(X,Y,\delta)\sim\mathbb{Q}}[\widetilde{\ell}(\theta;X,Y,\delta)].
\label{eq:dro}
\end{equation}
This is the worst-case expected loss when we sample from any distribution in $\mathcal{B}_{r}(\mathbb{P})$.

The definition for $\mathcal{B}_{r}(\mathbb{P})$ is somewhat technical; we first give its precise definition and then state how to choose $r$ so that $\mathcal{R}_{\text{DRO}}(\theta;r)$ is an upper bound on $\mathcal{R}_{\max}(\theta)$. Importantly, we will be able to efficiently minimize an empirical version of $\mathcal{R}_{\text{DRO}}(\theta;r)$.

\vspace{-2pt}
\begin{definition}
The set $\mathcal{B}_{r}(\mathbb{P})$ consists of all distributions $\mathbb{Q}$ that have the same (or smaller) support as $\mathbb{P}$ and have $\chi^{2}$-divergence is at most $r$ from distribution $\mathbb{P}$. Formally,
\[
\mathcal{B}_{r}(\mathbb{P}):=\{\text{dist.~}\mathbb{Q}\mid\mathbb{Q}\ll\mathbb{P}, D_{\chi^{2}}(\mathbb{Q}\|\mathbb{P})\le r\},
\]
where the notation ``~\!$\mathbb{Q}\ll\mathbb{P}$'' roughly means that $\mathbb{Q}$ has the same (or smaller) support as $\mathbb{P}$.\footnote{The measure-theoretic definition of ``$\mathbb{Q}\ll\mathbb{P}$'' is that $\mathbb{Q}$ is absolutely continuous with respect to~$\mathbb{P}$.} Meanwhile, $D_{\chi^2}(\mathbb{Q}\|\mathbb{P}):=\int (\frac{d \mathbb{Q}}{d \mathbb{P}} - 1)^2 d\mathbb{P}$.
\end{definition} \vspace{-2pt}
Working with $\mathcal{B}_{r}(\mathbb{P})$ turns out to be straightforward so long as we have a lower bound on the smallest group's probability (i.e., a lower bound on $\min_{k=1,\dots,K}\pi_k$). %

\vspace{-2pt}\vspace{-2pt}
\begin{proposition}(Directly follows from Proposition 2 of \citet{hashimoto2018fairness})
Suppose that we have a lower bound $\alphamin>0$ on the $K$ latent groups' probabilities of occurring (i.e., $\alphamin\le\min_{k=1,\dots,K}\pi_{k}$). Then $\mathcal{R}_{\text{DRO}}(\theta;r_{\max})\ge\mathcal{R}_{\max}(\theta)$, where $r_{\max}:=(\frac{1}{\alphamin}-1)^{2}$.
\end{proposition} \vspace{-2pt}\vspace{-2pt}
In other words, if we have a guess for $\alphamin\in(0,\min_{k=1,\dots,K}\pi_{k}]$, then it suffices to choose $r$ for $\mathcal{B}_{r}(\mathbb{P})$ to be $r_{\max}=(\frac{1}{\alphamin}-1)^{2}$. Furthermore, the risk $\mathcal{R}_{\text{DRO}}(\theta;r_{\max})$ is an upper bound on $\mathcal{R}_{\max}(\theta)$. In practice, $\alphamin\in(0,1)$ is a user-specified hyperparameter since we do not know $\pi_1,\dots,\pi_K$ nor $K$. Choosing $\alphamin$ to be smaller means that we want to ensure that groups with smaller probabilities of occurring also have low expected loss. For example, setting $\alpha=0.1$ means that the ``rarest'' group that we want to ensure low expected loss for occurs with probability least~0.1.

\vspace{-.5em}
\subsection{Empirical DRO Risk}
\label{sec:empirical-dro-risk-minimization}
\vspace{-.25em}

The next issue is how to minimize the risk $\mathcal{R}_{\text{DRO}}(\theta;r_{\max})$. This risk appears challenging to evaluate since it involves a supremum over all distributions in $\mathcal{B}_{r_{\max}}(\mathbb{P})$. However, a fundamental theoretical result from DRO literature is that $\mathcal{R}_{\text{DRO}}(\theta;r_{\max})$ can be written in a form that is amenable to computation.

\vspace{-2pt}\vspace{-2pt}\vspace{-2pt}
\begin{proposition} (Lemma 1 in \citet{duchi2021learning})
Suppose $\widehat{\ell}(\theta;X,Y,\delta)$ is upper semi-continuous with respect to $\theta$. Let $[\cdot]_+$ denote the ReLU function (i.e., $[a]_+:=\max\{a,0\}$ for any $a\in\mathbb{R}$), and $C:=\sqrt{2(\frac{1}{\alphamin} -1)^2+1}$. Then
\begin{align}
&\mathcal{R}_{\text{DRO}}(\theta;r_{\max})= \nonumber \\
&\!\inf_{\eta\in \mathbb{R}}\!\Big\{C\sqrt{\mathbb{E}_{(X,Y,\delta)\sim\mathbb{P}}\big[[\widetilde{\ell}(\theta;X,Y,\delta)-\eta]_+^2\big]}+\eta\Big\}.
\label{eq:dro_dual}
\end{align}
\end{proposition} \vspace{-2pt}\vspace{-2pt}\vspace{-2pt}\vspace{-2pt}
The right-hand side of equation~\eqref{eq:dro_dual} could be interpreted as follows. Suppose that we have achieved the optimal value $\eta^*$. Then the loss from a patient will be ignored if it is less than $\eta^*$ (due to the ReLU function). Thus, only the patients with losses above $\eta^*$ are considered for learning the survival model.

Note that as we vary the model parameters~$\theta$, the different patients' losses change. Thus, as a function of $\theta$, the DRO risk $\mathcal{R}_{\text{DRO}}(\theta;r_{\max})$ dynamically adjusts which patients to focus on, always prioritizing the patients with the highest loss values (again, we only consider the patients with a loss greater than the optimal value of $\eta$).

We can readily minimize an empirical version of $\mathcal{R}_{\text{DRO}}(\theta;r_{\max})$. Specifically, we replace the expectation on the right-hand side of equation~\eqref{eq:dro_dual} with an empirical average to arrive at the following optimization problem:
\begin{equation}
\min_{\theta\in\Theta,\eta\in\mathbb{R}}
\mathcal{L}_{\text{DRO}}(\theta, \eta),
\label{eq:empirical-dro-minimization}
\end{equation}
where $\Theta$ denotes the feasible set of the model parameters, and we define the empirical loss
\begin{align}
&\mathcal{L}_{\text{DRO}}(\theta, \eta) \nonumber \\
&:=C\sqrt{\frac{1}{n}\sum_{i=1}^n[\widetilde{\ell}(\theta;X_i,Y_i,\delta_i)-\eta]_+^2}+\eta.
\label{eq:empirical_dro}
\end{align}
\textbf{Numerical optimization.}
The optimization problem in equation~\eqref{eq:empirical-dro-minimization} can be solved with an iterative gradient descent approach \citep{hu2020learning, hu2021tkml, hu2022sum}. Specifically, we first initialize the model parameters $\theta$. Then, following \citet{hashimoto2018fairness}, we alternate between two steps:
\begin{itemize}[itemsep=0pt,topsep=2pt,parsep=0pt,leftmargin=*]
\item We fix $\theta$ and update $\eta$ by finding the value of $\eta$ that minimizes $\mathcal{L}_{\text{DRO}}(\theta, \eta)$. To do this, we use binary search to find the global optimum of $\eta$ since $\mathcal{L}_{\text{DRO}}(\theta, \eta)$ is a convex function with respect to $\eta$.
\item We fix $\eta$ and update $\theta$ by minimizing $\mathcal{L}_{\text{DRO}}(\theta, \eta)$ (e.g., using gradient descent).
\end{itemize}
We stop iterating after user-specified stopping criteria are reached (e.g., maximum number of iterations reached, early stopping due to no improvement in a validation metric after a pre-specified number of epochs). The pseudocode can be found in Appendix~\ref{sec:pseudocode}.

\vspace{-.5em}
\subsection{Choosing the Individual Loss \texorpdfstring{$\widetilde{\ell}$}{}}
\label{sec:individual-loss}
\vspace{-.25em}

The technical difficulty in applying DRO to the Cox model is somewhat subtle. In our description of DRO so far, the loss $\widetilde{\ell}$ that is mentioned depends only on model parameters $\theta$ and a single data point $(X,Y,\delta)$. In contrast, for the Cox model, the $i$-th patient's loss $\ell_{i}(\theta)$ as described in equation~\eqref{eq:cox-individual-loss} can actually depend on multiple training patients. The reason is that inside the log term of equation~\eqref{eq:cox-individual-loss}, there is a sum over all training patients $j=1,\dots,n$ whose observed time $Y_{j}$ is at least $Y_i$. Thus, replacing $\widetilde{\ell}(\theta;X_i,Y_i,\delta_i)$ in equation~\eqref{eq:empirical_dro} with the Cox individual loss $\ell_i(\theta)$ actually invalidates the theory we have covered thus far. However, our experiments later will reveal that this replacement works very well in practice. We call this method \textsc{dro-cox}. %

\smallskip
\noindent
\textbf{A theoretically sound DRO method for Cox models.}
We show how to define the individual loss $\widetilde{\ell}$ so that it complies with existing DRO theory. To achieve this, we use sample splitting and an approximation of the Cox individual loss. We divide the training patients into two sets $\mathcal{D}_1\subset\{1,\dots,n\}$ and $\mathcal{D}_2:=\{1,\dots,n\}\setminus\mathcal{D}_1$ of sizes $n_{1}:=|\mathcal{D}_1|$ and $n_{2}:=|\mathcal{D}_2|=n-n_{1}$. The high-level idea is that we only compute an approximation of the Cox individual loss $\ell_{i}(\theta)$ for $i\in\mathcal{D}_1$ (so that the empirical average in the DRO loss $\mathcal{L}_{\text{DRO}}(\theta,\eta)$ is modified to only be over the training patients in $\mathcal{D}_1$). Meanwhile, each $\ell_{i}(\theta)$ for $i\in\mathcal{D}_1$ is modified so that the sum inside the log term only depends on the $i$-th patient and patients in~$\mathcal{D}_2$.

In more detail, we approximate $\ell_{i}(\theta)$ for $i\in\mathcal{D}_1$ with the new individual loss
\begin{align}
& \widetilde{\ell}_{\text{split}}(\theta;X_i,Y_i,\delta_i,\mathcal{D}_2) \nonumber \\
& :=-\delta_i\big[f(X_i;\theta)-\log\big(\Phi(\theta;X_i,Y_i,\mathcal{D}_2)\big)\big], \label{eq:split_individual_loss}
\end{align}
where
\begin{align*}
 & \Phi(\theta;X_i,Y_i,\mathcal{D}_2)\\
 & :=\exp(f(X_i;\theta))+\!\!\!\!\sum_{j\in\mathcal{D}_2\text{ s.t.~}Y_{j}\ge Y_i}\!\!\!\!\exp(f(X_j;\theta)).
\end{align*}
This loss is no longer equal to the Cox individual loss $\ell_i(\theta)$ because the log term is computed only using the $i$-th training patient and patients in $\mathcal{D}_2$. Importantly, treating the training patients in $\mathcal{D}_2$ as fixed, then the individual loss $\widetilde{\ell}_{\text{split}}(\theta;X_i,Y_i,\delta_i,\mathcal{D}_2)$ only depends on $\theta$ and the data point $(X_i,Y_i,\delta_i)$. Note that our sample splitting strategy is somewhat inspired by the ``case control'' strategy by \citet{kvamme2019time}, where instead of using the full Cox loss, they approximate each individual data point's loss (which could depend on many other data points) to only depend on a single other data point.

We next modify the empirical DRO loss $\mathcal{L}_{\text{DRO}}(\theta, \eta)$ given in equation~\eqref{eq:empirical_dro} so that the empirical average (inside the square root) is only computed using training patients in $\mathcal{D}_1$, and we set $\widetilde{\ell}$ equal to $\widetilde{\ell}_{\text{split}}$. In particular, we replace $\mathcal{L}_{\text{DRO}}(\theta,\eta)$ with the loss
\begin{align}
&\mathcal{L}_{\text{DRO-split}}(\theta, \eta, \mathcal{D}_1, \mathcal{D}_2):= \nonumber \\
&~C\sqrt{\frac{1}{|\mathcal{D}_1|}\sum_{i\in\mathcal{D}_1}[\widetilde{\ell}_{\text{split}}(\theta;X_i,Y_i,\delta_i,\mathcal{D}_2)\!-\!\eta]_+^2}\!+\!\eta. \nonumber \\[-1em]
\label{eq:empirical_dro_split}
\end{align}
Although minimizing $\mathcal{L}_{\text{DRO-split}}(\theta, \eta, \mathcal{D}_1, \mathcal{D}_2)$ is compliant with DRO theory, it uses data less effectively since at most $n_1$ patients (rather than $n$) are used to compute the empirical average (note that only uncensored patients have nonzero loss), and for these patients, at most $n_2+1$ points are used to compute the sum inside each of their log terms.

A simple way to more effectively use the data is to change optimization problem~\eqref{eq:empirical-dro-minimization} to instead minimize the sum of two losses: the first is $\mathcal{L}_{\text{DRO-split}}(\theta, \eta, \mathcal{D}_1, \mathcal{D}_2)$, and the second is $\mathcal{L}_{\text{DRO-split}}(\theta, \eta', \mathcal{D}_2, \mathcal{D}_1)$, i.e., the latter loss swaps the roles of $\mathcal{D}_1$ and $\mathcal{D}_2$ and also $\eta$ is replaced with a different $\eta'$ (the two losses do not share the same $\eta$). The iterative optimization procedure in Section~\ref{sec:empirical-dro-risk-minimization} can still be applied except where each iteration now consists of three steps: updating $\eta$, $\eta'$, and $\theta$. We refer to this method as \mbox{\textsc{dro-cox (split)}}; we provide pseudocode for it in Appendix~\ref{sec:pseudocode}.

\vspace{-1.1em}
\section{Experiments}
\vspace{-.3em}

We compare \textsc{dro-cox} and \textsc{dro-cox (split)} against various baselines using a similar experimental setup as \citet{keya2021equitable}.

\smallskip
\noindent
\textbf{Datasets.}
We use three standard, publicly available survival analysis datasets that have been used in fair survival analysis research:
\begin{itemize}[itemsep=0pt,topsep=1pt,parsep=1pt,leftmargin=*]

\item The \textbf{FLC} dataset \citep{dispenzieri2012use} is from a study on the relationship between serum free light chain (FLC) and mortality of Olmsted County residents aged 50 or higher. We regard binary encoded age (age$\leq$65 and age$>$65) and gender (women and men) as sensitive attributes.

\item The \textbf{SUPPORT} dataset \citep{knaus1995support} is from a study at Vanderbilt University on understanding prognoses, preferences, outcomes, and risks of treatment by analyzing survival times of severely ill hospitalized patients. We regard binary encoded age (age$\leq$65 and age$>$65), race (white and non-white), and gender (women and men) as sensitive attributes.

\item The \textbf{SEER} dataset \citep{Teng2019SEER} of breast cancer patients is obtained from the 2017 November update of the Sureillance, Epidemiology, and End Results (SEER) program of the National Cancer Institute. The dataset is on female patients with %
breast cancer diagnosed in 2006-2010. We regard binary encoded age (age$\leq$65 and age$>$65) and race (white and non-white) as sensitive attributes.

\end{itemize} %
Basic characteristics of these datasets are reported in Table \ref{tab:overview_datasets}. For all datasets, we first use a random 80\%/20\% train/test split to hold out a test set that will be the same across experimental repeats. Then we repeat the following basic experiment 10 times: (1) We hold out 20\% of the training data to treat as a validation set, which is used to tune hyperparameters. (2) We then compute evaluation metrics across the same test set. We describe the evaluation metrics and how hyperparameter tuning works shortly. When we report our experimental results, we provide the mean and standard deviation of each metric across the 10 experimental repeats.

\begin{table}[t]
\vspace{-.6em}
\caption{Basic dataset characteristics.}\vspace{.5em}
\label{tab:overview_datasets}
\centering
\setlength\tabcolsep{2.5pt}
\scriptsize
{\renewcommand{\arraystretch}{1.3}
\begin{tabular}{cccc}
\toprule
& FLC & SUPPORT & SEER \\ \midrule
\# samples & 7,874 & 9,105 & 4,024 \\ %
\# features & 6 (9$^*$) & 14 (19$^*$) & 13 \\ %
Censoring rate & 0.725 & 0.319 & 0.847 \\ %
\makecell{Sensitive \\ attributes} & age, gender & age, race, gender & age, race
\\ \bottomrule
\end{tabular}
} \\[3pt]
$^*$ indicates the number before preprocessing \\ (preprocessing removes some features)
\vspace{-1.7em}
\end{table}

\smallskip
\noindent
\textbf{Evaluation metrics.}
We use accuracy and, separately, fairness metrics. The accuracy metrics we use are (a) concordance index (abbreviated as ``c-index'', higher is better) \citep{harrell1982evaluating}, (b) time-dependent AUC (AUC, higher is better) \citep{chambless2006estimation}, (c) log partial likelihood (LPL, higher is better), and (d) integrated IPCW Brier Score (IBS, lower is better) \citep{graf1999assessment}. Note that the \emph{negative} LPL averaged across data points is precisely given by equation~\eqref{eq:average} (all methods we consider are variants of Cox models).

As our experimental setup is largely based on that of \citet{keya2021equitable}, we use the fairness metrics that they had defined: individual fairness (F$_I$), group fairness (F$_G$), and intersectional fairness (F$_{\cap}$). We also include a summary fairness metric
$\text{F}_A=(\textrm{F}_I+\textrm{F}_G+\textrm{F}_{\cap})/3$.
As we pointed out in Section~\ref{sec:related_work}, the fairness metrics by \citet{keya2021equitable} do not actually account for accuracy. We thus also include the concordance imparity (CI) fairness metric by \citet{zhang2022longitudinal} that is based on accuracy. For all fairness metrics, lower is better. Definitions of these fairness metrics are in Appendix~\ref{sec:fairness-measures}.

Note that the fairness metrics F$_G$ and CI require us to specify groups. For the FLC dataset, we separately use (binary encoded) age and gender (i.e., we first run experiments using only age in evaluating F$_G$ and CI; we then re-run experiments using gender instead of age). For the SUPPORT dataset, we separately use gender, age, and race. For the SEER dataset, we separately use race and age. Note that since F$_A$ depends on F$_G$, the F$_A$ metric also changes when we switch the sensitive attribute used for F$_G$ and CI. Meanwhile, the intersectional fairness metric F$_{\cap}$ is meant for when multiple sensitive attributes are specified. Per dataset, we use all sensitive attributes specified in Table~\ref{tab:overview_datasets} to evaluate~F$_{\cap}$.

\smallskip
\noindent
\textbf{Methods evaluated.}
For simplicity, all models evaluated are Cox models, either assuming the \textbf{linear} setting (the log partial hazard function is $f(x;\theta)=\theta^T x$) or the \textbf{nonlinear} setting in which $f$ is a multilayer perceptron (MLP). When \textsc{dro-cox} or \textsc{dro-cox (split)} are used in the latter case, we add the prefix ``Deep'' in tables for clarity.

For baselines, the unregularized linear Cox model \citep{cox1972regression} is denoted as ``Cox'' in our tables, whereas the unregularized nonlinear Cox model \citep{katzman2018deepsurv} is denoted as ``DeepSurv''. The rest of our baselines are all regularized versions of either the standard Cox or DeepSurv models, using different fairness regularization terms. When we use individual, group, or intersectional regularization terms by \citet{keya2021equitable}, then we add the suffix ``$_I$(Keya~et~al.)'', ``$_G$(Keya~et~al.)'', or ``$_{\cap}$(Keya~et~al.)'' respectively to a model name; for example, ``DeepSurv$_G$(Keya~et~al.)'' corresponds to DeepSurv with group fairness regularization by \citet{keya2021equitable}. When we use the individual or group fairness regularization terms that account for observed times and censoring information \citep{rahman2022fair}, we instead use the suffix ``$_I$(R\&P)'' or ``$_G$(R\&P)''.\footnote{\citet{rahman2022fair} did not propose an intersectional fairness regularizer and technically did not try regularized versions of Cox models using their fairness definitions. However, it is straightforward to adapt their individual and group fairness definitions as regularization terms for a Cox model, especially as their work is directly modifying definitions by \citet{keya2021equitable}.} Note that group fairness regularization (suffixes ``$_G$(Keya et al.)'' and ``$_G$(R\&P)'') uses the same groups that test set F$_G$ and CI fairness metrics use.

Hyperparameter grids for all methods (including our \textsc{dro-cox} variants) are in Appendix~\ref{sec:hyperparameters-compute-env}, where we also provide information on the compute environment that we used. In terms of hyperparameter tuning, we use the strategy by \citet{keya2021equitable}: the final hyperparameter setting used per dataset and per method is determined based on a preset rule in practice that allows up to a 5\% degradation in the validation set c-index from the classical Cox model (for the linear setting) or DeepSurv (for the nonlinear setting) while minimizing the validation set CI fairness metric (Keya et al.~used their own fairness metrics though instead of CI).

\begin{table*}[pth!]
\vspace{-1em}
\caption{\small Test set accuracy and fairness metrics on the FLC (age) dataset. We report mean and standard deviation (in parentheses) across 10 experimental repeats (each repeat holds out a different 20\% of the training data as a validation set for hyperparameter tuning; the test set is the same across experimental repeats). Higher is better for metrics with ``$\uparrow$'', while lower is better for metrics with ``$\downarrow$''. The best results are shown in bold for linear and, separately, nonlinear models. When one of our methods outperforms all baselines (in linear and, separately, nonlinear models), we highlight the corresponding cell in \mybox[fill=green!30]{green}.} \vspace{-1em}
\centering
\setlength\tabcolsep{3.6pt}
{\tiny %
\renewcommand{\belowrulesep}{0.1pt}
\renewcommand{\aboverulesep}{0.1pt}
\begin{tabular}{cccccc|ccccc}
\toprule
\multirow{2}{*}{} & \multirow{2}{*}{Methods} & \multicolumn{4}{c|}{Accuracy Metrics}                                                    & \multicolumn{5}{c}{Fairness Metrics}                             \\ \cmidrule{3-11}
&                   & \multicolumn{1}{c}{c-index$\uparrow$} & \multicolumn{1}{c}{AUC$\uparrow$} & \multicolumn{1}{c}{LPL$\uparrow$} & IBS$\downarrow$ & \multicolumn{1}{c}{F$_I$$\downarrow$} & \multicolumn{1}{c}{F$_G$$\downarrow$} & \multicolumn{1}{c}{F$_{\cap}$$\downarrow$}& \multicolumn{1}{c}{F$_A$$\downarrow$} &CI(\%)$\downarrow$                  \\ \midrule
\multirow{8}{*}{\rotatebox{90}{Linear \ \ \ \ \ \ \ \ \ \ \ \ \ \ }} &        \tableCox            & \multicolumn{1}{c}{{\makecell{0.8032 \\(0.0002)}}} & \multicolumn{1}{c}{\makecell{0.8176 \\(0.0005)}} & \multicolumn{1}{c}{{\textbf{\makecell{-6.3724 \\(0.0011)}}}} & {\makecell{0.1739 \\(0.0004)}} & \multicolumn{1}{c}{\makecell{1.8787 \\(0.0304)}} & \multicolumn{1}{c}{\makecell{3.0282 \\(0.0469)}} & \multicolumn{1}{c}{\makecell{2.8355 \\(0.0297)}} & \multicolumn{1}{c}{\makecell{2.5808 \\(0.0332)}} & \makecell{0.5350 \\(0.0413)}              \\ \cmidrule{3-11}
&          \tableCoxKeyaInd          & \multicolumn{1}{c}{{\makecell{0.7937 \\(0.0068)}}} & \multicolumn{1}{c}{\makecell{0.8179 \\(0.0067)}} & \multicolumn{1}{c}{{\makecell{-6.6044 \\(0.0721)}}} & {\makecell{0.1414 \\(0.0073)}} & \multicolumn{1}{c}{\makecell{0.4493 \\(0.1217)}} & \multicolumn{1}{c}{\makecell{0.7623 \\(0.1995)}} & \multicolumn{1}{c}{\makecell{1.2045 \\(0.2701)}} & \multicolumn{1}{c}{\makecell{0.8054 \\(0.1966)}} & \makecell{0.5400 \\(0.3270)}                 \\ \cmidrule{3-11}
&          \tableCoxRPInd          & \multicolumn{1}{c}{{\textbf{\makecell{0.8034 \\(0.0007)}}}} & \multicolumn{1}{c}{\makecell{0.8188 \\(0.0008)}} & \multicolumn{1}{c}{{\makecell{-6.4111 \\(0.0203)}}} & {\makecell{0.1636 \\(0.0030)}} & \multicolumn{1}{c}{\makecell{1.0920 \\(0.1391)}} & \multicolumn{1}{c}{\makecell{1.7990 \\(0.2242)}} & \multicolumn{1}{c}{\makecell{2.1828 \\(0.1620)}} & \multicolumn{1}{c}{\makecell{1.6913 \\(0.1747)}} & \makecell{0.4330 \\(0.1196)}                 \\ \cmidrule{3-11}
&        \tableCoxKeyaGroup            & \multicolumn{1}{c}{{\makecell{0.7974 \\(0.0117)}}} & \multicolumn{1}{c}{\textbf{\makecell{0.8196 \\(0.0063)}}} & \multicolumn{1}{c}{{\makecell{-6.6869 \\(0.0693)}}} & {\makecell{0.1492 \\(0.0077)}} & \multicolumn{1}{c}{\makecell{1.0495 \\(0.6647)}} & \multicolumn{1}{c}{\makecell{1.1802 \\(0.6893)}} & \multicolumn{1}{c}{\makecell{1.7940 \\(0.7234)}} & \multicolumn{1}{c}{\makecell{1.3412 \\(0.6880)}} & \makecell{0.3410 \\(0.3011)}                 \\ \cmidrule{3-11}
&        \tableCoxRPGroup            & \multicolumn{1}{c}{{\makecell{0.8027 \\(0.0005)}}} & \multicolumn{1}{c}{\makecell{0.8172 \\(0.0011)}} & \multicolumn{1}{c}{{\makecell{-6.3921 \\(0.0095)}}} & {\makecell{0.1676 \\(0.0012)}} & \multicolumn{1}{c}{\makecell{1.3130 \\(0.0801)}} & \multicolumn{1}{c}{\makecell{2.1601 \\(0.1296)}} & \multicolumn{1}{c}{\makecell{2.3984 \\(0.0903)}} & \multicolumn{1}{c}{\makecell{1.9571 \\(0.0992)}} & \makecell{0.4950 \\(0.1517)}                 \\ \cmidrule{3-11}
&        \tableCoxKeyaInt            & \multicolumn{1}{c}{{\makecell{0.7870 \\(0.0029)}}} & \multicolumn{1}{c}{\makecell{0.8148 \\(0.0017)}} & \multicolumn{1}{c}{{\makecell{-6.7272 \\(0.0048)}}} & {\textbf{\makecell{0.1400 \\(0.0005)}}} & \multicolumn{1}{c}{\makecell{0.2921 \\(0.0056)}} & \multicolumn{1}{c}{\textbf{\makecell{0.4156 \\(0.0220)}}} & \multicolumn{1}{c}{\textbf{\makecell{0.6827 \\(0.0291)}}} & \multicolumn{1}{c}{\textbf{\makecell{0.4635 \\(0.0180)}}} & \makecell{1.0790 \\(0.1098)}                 \\ \cmidrule{2-11} 
&       \tableDROCox             & \multicolumn{1}{c}{{\makecell{0.7959 \\(0.0036)}}} & \multicolumn{1}{c}{\makecell{0.8149 \\(0.0020)}} & \multicolumn{1}{c}{{\makecell{-6.7630 \\(0.2113)}}} & {\makecell{0.1408 \\(0.0050)}} & \multicolumn{1}{c}{\1 \textbf{\makecell{0.2793 \\(0.1818)}}} & \multicolumn{1}{c}{\makecell{0.4694 \\(0.3016)}} & \multicolumn{1}{c}{\makecell{0.7880 \\(0.5011)}} & \multicolumn{1}{c}{\makecell{0.5122 \\(0.3282)}} & \1\textbf{\makecell{0.0510 \\(0.0401)}}                \\ \cmidrule{3-11}
&       \tableDROCoxSplit            & \multicolumn{1}{c}{{\makecell{0.8017 \\(0.0017)}}} & \multicolumn{1}{c}{\makecell{0.8221 \\(0.0014)}} & \multicolumn{1}{c}{{\makecell{-6.6593 \\(0.2604)}}} & {\makecell{0.1658 \\(0.0098)}} & \multicolumn{1}{c}{\makecell{0.5611 \\(0.3588)}} & \multicolumn{1}{c}{\makecell{0.8935 \\(0.5652)}} & \multicolumn{1}{c}{\makecell{1.2734 \\(0.7880)}} & \multicolumn{1}{c}{\makecell{0.9094 \\(0.5705)}} & \1\makecell{0.0840 \\(0.0594)}                 \\ \cmidrule{1-11}
\multirow{8}{*}{\rotatebox{90}{Nonlinear \ \ \ \ \ \ \ \ \ \ \ \ \ \ \  }} &          DeepSurv         & \multicolumn{1}{c}{{\makecell{0.8070 \\(0.0014)}}} & \multicolumn{1}{c}{\makecell{0.8247 \\(0.0026)}} & \multicolumn{1}{c}{{\textbf{\makecell{-6.3552 \\(0.0052)}}}} & {\makecell{0.1767 \\(0.0018)}} & \multicolumn{1}{c}{\makecell{2.9691 \\(1.2481)}} & \multicolumn{1}{c}{\makecell{4.6647 \\(1.9185)}} & \multicolumn{1}{c}{\makecell{2.8800 \\(0.0531)}} & \multicolumn{1}{c}{\makecell{3.5046 \\(1.0506)}} & \makecell{0.2940 \\(0.2147)}                 \\  \cmidrule{3-11}
&          \tableDeepSurvKeyaInd          & \multicolumn{1}{c}{{\makecell{0.7884 \\(0.0070)}}} & \multicolumn{1}{c}{\makecell{0.8134 \\(0.0109)}} & \multicolumn{1}{c}{{\makecell{-6.6416 \\(0.1399)}}} & {\makecell{0.1441 \\(0.0130)}} & \multicolumn{1}{c}{\makecell{0.1510 \\(0.1062)}} & \multicolumn{1}{c}{\makecell{0.2425 \\(0.1675)}} & \multicolumn{1}{c}{\makecell{1.1281 \\(0.5625)}} & \multicolumn{1}{c}{\makecell{0.5072 \\(0.1335)}} & \makecell{0.3700 \\(0.2523)}                \\  \cmidrule{3-11}
&          \tableDeepSurvRPInd          & \multicolumn{1}{c}{{\makecell{0.8071 \\(0.0041)}}} & \multicolumn{1}{c}{\makecell{0.8254 \\(0.0049)}} & \multicolumn{1}{c}{{\makecell{-6.3824 \\(0.0743)}}} & {\makecell{0.1729 \\(0.0093)}} & \multicolumn{1}{c}{\textbf{\makecell{0.0713 \\(0.1204)}}} & \multicolumn{1}{c}{\makecell{0.1167 \\(0.1785)}} & \multicolumn{1}{c}{\makecell{2.5089 \\(0.4270)}} & \multicolumn{1}{c}{\makecell{0.8990 \\(0.0643)}} & \makecell{0.1870 \\(0.1117)}                \\  \cmidrule{3-11}
&        \tableDeepSurvKeyaGroup            & \multicolumn{1}{c}{{\makecell{0.7990 \\(0.0120)}}} & \multicolumn{1}{c}{\makecell{0.8189 \\(0.0108)}} & \multicolumn{1}{c}{{\makecell{-6.4954 \\(0.1924)}}} & {\makecell{0.4190 \\(0.2487)}} & \multicolumn{1}{c}{\makecell{0.1604 \\(0.3575)}} & \multicolumn{1}{c}{\textbf{\makecell{0.1600 \\(0.3249)}}} & \multicolumn{1}{c}{\makecell{1.0645 \\(0.6657)}} & \multicolumn{1}{c}{\makecell{0.4617 \\(0.4381)}} & \makecell{0.2490 \\(0.1646)}              \\ \cmidrule{3-11}
&        \tableDeepSurvRPGroup            & \multicolumn{1}{c}{{\textbf{\makecell{0.8073 \\(0.0036)}}}} & \multicolumn{1}{c}{\makecell{0.8255 \\(0.0049)}} & \multicolumn{1}{c}{{\makecell{-6.3786 \\(0.0687)}}} & {\makecell{0.1731 \\(0.0087)}} & \multicolumn{1}{c}{\makecell{0.2376 \\(0.2349)}} & \multicolumn{1}{c}{\makecell{0.3749 \\(0.3587)}} & \multicolumn{1}{c}{\makecell{2.6416 \\(0.4063)}} & \multicolumn{1}{c}{\makecell{1.0847 \\(0.1954)}} & \makecell{0.2290 \\(0.1344)}               \\ \cmidrule{3-11}
&          \tableDeepSurvKeyaInt          & \multicolumn{1}{c}{{\makecell{0.7751 \\(0.0018)}}} & \multicolumn{1}{c}{\makecell{0.7893 \\(0.0022)}} & \multicolumn{1}{c}{{\makecell{-6.8458 \\(0.0031)}}} & {\textbf{\makecell{0.1357 \\(0.0002)}}} & \multicolumn{1}{c}{\makecell{0.1688 \\(0.0035)}} & \multicolumn{1}{c}{\makecell{0.2412 \\(0.0051)}} & \multicolumn{1}{c}{\textbf{\makecell{0.4633 \\(0.0106)}}} & \multicolumn{1}{c}{\textbf{\makecell{0.2911 \\(0.0062}}} & \makecell{0.4300 \\(0.1091)}                 \\ \cmidrule{2-11} 
&          \tableDeepDROCox          & \multicolumn{1}{c}{{\makecell{0.8068 \\(0.0024)}}} & \multicolumn{1}{c}{\1\textbf{\makecell{0.8259 \\(0.0031)}}} & \multicolumn{1}{c}{{\makecell{-6.4698 \\(0.1069)}}} & {\makecell{0.1595 \\(0.0135)}} & \multicolumn{1}{c}{\makecell{1.3709 \\(1.1919)}} & \multicolumn{1}{c}{\makecell{2.1481 \\(1.8343)}} & \multicolumn{1}{c}{\makecell{1.8712 \\(0.6223)}} & \multicolumn{1}{c}{\makecell{1.7967 \\(1.1697)}} & \1\textbf{\makecell{0.0730 \\(0.0822)}}                 \\ \cmidrule{3-11}
&       \tableDeepDROCoxSplit            & \multicolumn{1}{c}{{\makecell{0.7650 \\(0.0024)}}} & \multicolumn{1}{c}{\makecell{0.7744 \\(0.0022)}} & \multicolumn{1}{c}{{\makecell{-6.8071 \\(0.0091)}}} & {\makecell{0.1703 \\(0.0002)}} & \multicolumn{1}{c}{\makecell{0.4480 \\(0.1050)}} & \multicolumn{1}{c}{\makecell{0.5327 \\(0.0706)}} & \multicolumn{1}{c}{\makecell{0.7762 \\(0.0992)}} & \multicolumn{1}{c}{\makecell{0.5856 \\(0.0914)}} & \makecell{2.8000 \\(0.1450)}                 \\ \bottomrule
\end{tabular}
}
\label{tab:general_performance_CI}
\vspace{-1em}
\end{table*}

\begin{table*}[pth!]
\caption{\small Test set scores on the SUPPORT (gender) dataset, in the same format as Table~\ref{tab:general_performance_CI}.}
\centering
\setlength\tabcolsep{3.6pt}
{\tiny %
\renewcommand{\belowrulesep}{0.1pt}
\renewcommand{\aboverulesep}{0.1pt}
\begin{tabular}{cccccc|ccccc}
\toprule
\multirow{2}{*}{} & \multirow{2}{*}{Methods} & \multicolumn{4}{c|}{Accuracy Metrics}                                                    & \multicolumn{5}{c}{Fairness Metrics}                             \\ \cmidrule{3-11}
&                   & \multicolumn{1}{c}{c-index$\uparrow$} & \multicolumn{1}{c}{AUC$\uparrow$} & \multicolumn{1}{c}{LPL$\uparrow$} & IBS$\downarrow$ & \multicolumn{1}{c}{F$_I$$\downarrow$} & \multicolumn{1}{c}{F$_G$$\downarrow$} & \multicolumn{1}{c}{F$_{\cap}$$\downarrow$}& \multicolumn{1}{c}{F$_A$$\downarrow$} &CI(\%)$\downarrow$                  \\ \midrule
\multirow{8}{*}{\rotatebox{90}{Linear \ \ \ \ \ \ \ \ \ \ \ \ \ \ }} &        \tableCox            & \multicolumn{1}{c}{{\makecell{0.6025 \\(0.0005)}}} & \multicolumn{1}{c}{\makecell{0.6163 \\(0.0010)}} & \multicolumn{1}{c}{{\textbf{\makecell{-6.8761 \\(0.0010)}}}} & {\makecell{0.2304 \\(0.0015)}} & \multicolumn{1}{c}{\makecell{0.2113 \\(0.0093)}} & \multicolumn{1}{c}{\makecell{0.0439 \\(0.0052)}} & \multicolumn{1}{c}{\makecell{0.4490 \\(0.0322)}} & \multicolumn{1}{c}{\makecell{0.2347 \\(0.0127)}} & \makecell{1.4300 \\(0.0654)}               \\  \cmidrule{3-11}
&          \tableCoxKeyaInd          & \multicolumn{1}{c}{{\makecell{0.5881 \\(0.0114)}}} & \multicolumn{1}{c}{\makecell{0.5998 \\(0.0142)}} & \multicolumn{1}{c}{{\makecell{-6.9387 \\(0.0202)}}} & {\textbf{\makecell{0.2157 \\(0.0060)}}} & \multicolumn{1}{c}{\makecell{0.0382 \\(0.0320)}} & \multicolumn{1}{c}{\makecell{0.0076 \\(0.0057)}} & \multicolumn{1}{c}{\makecell{0.0938 \\(0.0700)}} & \multicolumn{1}{c}{\makecell{0.0465 \\(0.0353)}} & \makecell{0.9650 \\(0.6126)}                  \\ \cmidrule{3-11}
&          \tableCoxRPInd          & \multicolumn{1}{c}{{\makecell{0.6019 \\(0.0019)}}} & \multicolumn{1}{c}{\makecell{0.6159 \\(0.0029)}} & \multicolumn{1}{c}{{\makecell{-6.8798 \\(0.0029)}}} & {\makecell{0.2282 \\(0.0013)}} & \multicolumn{1}{c}{\makecell{0.1814 \\(0.0127)}} & \multicolumn{1}{c}{\makecell{0.0383 \\(0.0139)}} & \multicolumn{1}{c}{\makecell{0.3841 \\(0.0240)}} & \multicolumn{1}{c}{\makecell{0.2013 \\(0.0126)}} & \makecell{1.4190 \\(0.1002)}                  \\ \cmidrule{3-11}
&        \tableCoxKeyaGroup            & \multicolumn{1}{c}{{\textbf{\makecell{0.6030 \\(0.0007)}}}} & \multicolumn{1}{c}{\textbf{\makecell{0.6177 \\(0.0011)}}} & \multicolumn{1}{c}{{\makecell{-6.8772 \\(0.0017)}}} & {\makecell{0.2297 \\(0.0018)}} & \multicolumn{1}{c}{\makecell{0.2016 \\(0.0117)}} & \multicolumn{1}{c}{\makecell{0.0032 \\(0.0016)}} & \multicolumn{1}{c}{\makecell{0.4012 \\(0.0163)}} & \multicolumn{1}{c}{\makecell{0.2020 \\(0.0071)}} & \makecell{1.4190 \\(0.0632)}                 \\ \cmidrule{3-11}
&        \tableCoxRPGroup            & \multicolumn{1}{c}{{\makecell{0.6018 \\(0.0017)}}} & \multicolumn{1}{c}{\makecell{0.6156 \\(0.0027)}} & \multicolumn{1}{c}{{\makecell{-6.8779 \\(0.0023)}}} & {\makecell{0.2295 \\(0.0009)}} & \multicolumn{1}{c}{\makecell{0.1993 \\(0.0048)}} & \multicolumn{1}{c}{\makecell{0.0430 \\(0.0157)}} & \multicolumn{1}{c}{\makecell{0.4239 \\(0.0281)}} & \multicolumn{1}{c}{\makecell{0.2221 \\(0.0128)}} & \makecell{1.4340 \\(0.1039)}                 \\ \cmidrule{3-11}
&        \tableCoxKeyaInt            & \multicolumn{1}{c}{{\makecell{0.5715 \\(0.0062)}}} & \multicolumn{1}{c}{\makecell{0.5718 \\(0.0081)}} & \multicolumn{1}{c}{{\makecell{-6.9078 \\(0.0062)}}} & {\makecell{0.2275 \\(0.0016)}} & \multicolumn{1}{c}{\makecell{0.1334 \\(0.0129)}} & \multicolumn{1}{c}{\makecell{0.0092 \\(0.0037)}} & \multicolumn{1}{c}{\makecell{0.0743 \\(0.0108)}} & \multicolumn{1}{c}{\makecell{0.0723 \\(0.0077)}} & \makecell{1.1270 \\(0.2457)}               \\ \cmidrule{2-11} 
&       \tableDROCox          & \multicolumn{1}{c}{{\makecell{0.5734 \\(0.0019)}}} & \multicolumn{1}{c}{\makecell{0.6083 \\(0.0023)}} & \multicolumn{1}{c}{{\makecell{-6.9388 \\(0.0007)}}} & {\makecell{0.2210 \\(0.0010)}} & \multicolumn{1}{c}{\1\makecell{0.0378 \\(0.0013)}} & \multicolumn{1}{c}{\1\textbf{\makecell{0.0022 \\(0.0015)}}} & \multicolumn{1}{c}{\1\makecell{0.0731 \\(0.0094)}} & \multicolumn{1}{c}{\1\makecell{0.0377 \\(0.0033)}} & \1\makecell{0.4350 \\(0.0674)}            \\ \cmidrule{3-11}
&       \tableDROCoxSplit          & \multicolumn{1}{c}{{\makecell{0.5725 \\(0.0075)}}} & \multicolumn{1}{c}{\makecell{0.6056 \\(0.0092)}} & \multicolumn{1}{c}{{\makecell{-6.9410 \\(0.0057)}}} & {\makecell{0.4264 \\(0.1667)}} & \multicolumn{1}{c}{\1\textbf{\makecell{0.0285 \\(0.0188)}}} & \multicolumn{1}{c}{\makecell{0.0041 \\(0.0028)}} & \multicolumn{1}{c}{\1\textbf{\makecell{0.0594 \\(0.0366)}}} & \multicolumn{1}{c}{\1\textbf{\makecell{0.0307 \\(0.0190)}}} & \1\textbf{\makecell{0.3410 \\(0.1781)}}                 \\ \cmidrule{1-11} 
\multirow{8}{*}{\rotatebox{90}{Nonlinear \ \ \ \ \ \ \ \ \ \ \ \ \ \ \   }} &          DeepSurv         & \multicolumn{1}{c}{{\makecell{0.6108 \\(0.0029)}}} & \multicolumn{1}{c}{\textbf{\makecell{0.6327 \\(0.0045)}}} & \multicolumn{1}{c}{{\textbf{\makecell{-6.8754 \\(0.0040)}}}} & {\makecell{0.2417 \\(0.0016)}} & \multicolumn{1}{c}{\makecell{0.4072 \\(0.0369)}} & \multicolumn{1}{c}{\makecell{0.0570 \\(0.0180)}} & \multicolumn{1}{c}{\makecell{0.4244 \\(0.0573)}} & \multicolumn{1}{c}{\makecell{0.2962 \\(0.0283)}} & \makecell{1.6220 \\(0.3303)}              \\  \cmidrule{3-11}
&          \tableDeepSurvKeyaInd          & \multicolumn{1}{c}{{\makecell{0.5984 \\(0.0124)}}} & \multicolumn{1}{c}{\makecell{0.6164 \\(0.0150)}} & \multicolumn{1}{c}{{\makecell{-6.9130 \\(0.0220)}}} & {\makecell{0.2376 \\(0.0182)}} & \multicolumn{1}{c}{\textbf{\makecell{0.0179 \\(0.0249)}}} & \multicolumn{1}{c}{\makecell{0.0059 \\(0.0076)}} & \multicolumn{1}{c}{\makecell{0.5688 \\(0.3030)}} & \multicolumn{1}{c}{\makecell{0.1975 \\(0.0930)}} & \makecell{1.3280 \\(0.7670)}                \\  \cmidrule{3-11}
&          \tableDeepSurvRPInd          & \multicolumn{1}{c}{{\makecell{0.6104 \\(0.0076)}}} & \multicolumn{1}{c}{\makecell{0.6315 \\(0.0115)}} & \multicolumn{1}{c}{{\makecell{-6.8761 \\(0.0132)}}} & {\makecell{0.2379 \\(0.0079)}} & \multicolumn{1}{c}{\makecell{0.0544 \\(0.0468)}} & \multicolumn{1}{c}{\makecell{0.0132 \\(0.0141)}} & \multicolumn{1}{c}{\makecell{0.4997 \\(0.1722)}} & \multicolumn{1}{c}{\makecell{0.1891 \\(0.0435)}} & \makecell{1.6490 \\(0.2368)}               \\  \cmidrule{3-11}
&        \tableDeepSurvKeyaGroup           & \multicolumn{1}{c}{{\makecell{0.5982 \\(0.0109)}}} & \multicolumn{1}{c}{\makecell{0.6176 \\(0.0144)}} & \multicolumn{1}{c}{{\makecell{-6.9121 \\(0.0278)}}} & {\makecell{0.2436 \\(0.0121)}} & \multicolumn{1}{c}{\makecell{0.1131 \\(0.0718)}} & \multicolumn{1}{c}{\textbf{\makecell{0.0047 \\(0.0036)}}} & \multicolumn{1}{c}{\makecell{0.3972 \\(0.1017)}} & \multicolumn{1}{c}{\makecell{0.1717 \\(0.0375)}} & \makecell{1.6540 \\(0.3892)}              \\ \cmidrule{3-11}
&        \tableDeepSurvRPGroup            & \multicolumn{1}{c}{{\textbf{\makecell{0.6110 \\(0.0057)}}}} & \multicolumn{1}{c}{\makecell{0.6325 \\(0.0089)}} & \multicolumn{1}{c}{{\makecell{-6.8766 \\(0.0117)}}} & {\makecell{0.2406 \\(0.0068)}} & \multicolumn{1}{c}{\makecell{0.0452 \\(0.0476)}} & \multicolumn{1}{c}{\makecell{0.0113 \\(0.0144}} & \multicolumn{1}{c}{\makecell{0.5246 \\(0.1217)}} & \multicolumn{1}{c}{\makecell{0.1937 \\(0.0320)}} & \makecell{1.6250 \\(0.1931)}             \\ \cmidrule{3-11}
&          \tableDeepSurvKeyaInt         & \multicolumn{1}{c}{{\makecell{0.6015 \\(0.0069)}}} & \multicolumn{1}{c}{\makecell{0.6190 \\(0.0100)}} & \multicolumn{1}{c}{{\makecell{-6.8794 \\(0.0055)}}} & {\makecell{0.2378 \\(0.0053)}} & \multicolumn{1}{c}{\makecell{0.2465 \\(0.0424)}} & \multicolumn{1}{c}{\makecell{0.0053 \\(0.0032)}} & \multicolumn{1}{c}{\textbf{\makecell{0.0745 \\(0.0263)}}} & \multicolumn{1}{c}{\makecell{0.1088 \\(0.0213)}} & \makecell{1.4110 \\(0.2129)}                \\ \cmidrule{2-11}  
&          \tableDeepDROCox          & \multicolumn{1}{c}{{\makecell{0.5829 \\(0.0067)}}} & \multicolumn{1}{c}{\makecell{0.6237 \\(0.0111)}} & \multicolumn{1}{c}{{\makecell{-6.9253 \\(0.0025)}}} & {\1\textbf{\makecell{0.2240 \\(0.0010)}}} & \multicolumn{1}{c}{\makecell{0.1109 \\(0.0377)}} & \multicolumn{1}{c}{\makecell{0.0058 \\(0.0021)}} & \multicolumn{1}{c}{\makecell{0.0816 \\(0.0095)}} & \multicolumn{1}{c}{\1\textbf{\makecell{0.0661 \\(0.0141)}}} & \1\textbf{\makecell{1.2600 \\(0.4412)}}             \\ \cmidrule{3-11}
&       \tableDeepDROCoxSplit            & \multicolumn{1}{c}{{\makecell{0.5448 \\(0.0015)}}} & \multicolumn{1}{c}{\makecell{0.5625 \\(0.0021)}} & \multicolumn{1}{c}{{\makecell{-6.9555 \\(0.0012)}}} & {\makecell{0.6390 \\(0.0005)}} & \multicolumn{1}{c}{\makecell{0.1605 \\(0.0030)}} & \multicolumn{1}{c}{\makecell{0.0071 \\(0.0024)}} & \multicolumn{1}{c}{\makecell{0.1754 \\(0.0062)}} & \multicolumn{1}{c}{\makecell{0.1143 \\(0.0031)}} & \makecell{2.1690 \\(0.0727)}              \\ \bottomrule
\end{tabular}
}
\label{tab:general_performance_SUPPORT_gender_CI}
\end{table*}

\begin{table*}[pth!]
\vspace{-1em}
\caption{\small Test set scores on the SEER (race) dataset, in the same format as Table~\ref{tab:general_performance_CI}.} 
\centering
\setlength\tabcolsep{3.6pt}
{\tiny %
\renewcommand{\belowrulesep}{0.1pt}
\renewcommand{\aboverulesep}{0.1pt}
\begin{tabular}{cccccc|ccccc}
\toprule
\multirow{2}{*}{} & \multirow{2}{*}{Methods} & \multicolumn{4}{c|}{Accuracy Metrics}                                                    & \multicolumn{5}{c}{Fairness Metrics}                             \\ \cmidrule{3-11}
&                   & \multicolumn{1}{c}{c-index$\uparrow$} & \multicolumn{1}{c}{AUC$\uparrow$} & \multicolumn{1}{c}{LPL$\uparrow$} & IBS$\downarrow$ & \multicolumn{1}{c}{F$_I$$\downarrow$} & \multicolumn{1}{c}{F$_G$$\downarrow$} & \multicolumn{1}{c}{F$_{\cap}$$\downarrow$}& \multicolumn{1}{c}{F$_A$$\downarrow$} &CI(\%)$\downarrow$                  \\ \midrule
\multirow{8}{*}{\rotatebox{90}{Linear \ \ \ \ \ \ \ \ \ \ \ \ \ \ }} &        \tableCox            & \multicolumn{1}{c}{{\textbf{\makecell{0.7409 \\(0.0016)}}}} & \multicolumn{1}{c}{\textbf{\makecell{0.7624 \\(0.0017)}}} & \multicolumn{1}{c}{{\textbf{\makecell{-5.9427 \\(0.0034)}}}} & {\makecell{0.0964 \\(0.0008)}} & \multicolumn{1}{c}{\makecell{0.6105 \\(0.0307)}} & \multicolumn{1}{c}{\makecell{0.1183 \\(0.0645)}} & \multicolumn{1}{c}{\makecell{0.6750 \\(0.0630)}} & \multicolumn{1}{c}{\makecell{0.4679 \\(0.0437)}} & \makecell{1.1880 \\(0.1742)}              \\  \cmidrule{3-11}
&          \tableCoxKeyaInd         & \multicolumn{1}{c}{{\makecell{0.7143 \\(0.0300)}}} & \multicolumn{1}{c}{\makecell{0.7367 \\(0.0303)}} & \multicolumn{1}{c}{{\makecell{-6.1602 \\(0.0728)}}} & {\makecell{0.0894 \\(0.0014)}} & \multicolumn{1}{c}{\makecell{0.1968 \\(0.0831)}} & \multicolumn{1}{c}{\makecell{0.0944 \\(0.0770)}} & \multicolumn{1}{c}{\makecell{0.3768 \\(0.1736)}} & \multicolumn{1}{c}{\makecell{0.2227 \\(0.0951)}} & \makecell{1.5500 \\(1.5117)}                 \\ \cmidrule{3-11}
&          \tableCoxRPInd         & \multicolumn{1}{c}{{\makecell{0.7353 \\(0.0195)}}} & \multicolumn{1}{c}{\makecell{0.7559 \\(0.0220)}} & \multicolumn{1}{c}{{\makecell{-6.0046 \\(0.0616)}}} & {\makecell{0.0919 \\(0.0012)}} & \multicolumn{1}{c}{\makecell{0.3655 \\(0.0720)}} & \multicolumn{1}{c}{\makecell{0.0707 \\(0.0567)}} & \multicolumn{1}{c}{\makecell{0.4031 \\(0.0698)}} & \multicolumn{1}{c}{\makecell{0.2798 \\(0.0396)}} & \makecell{1.5390 \\(0.5256)}                \\ \cmidrule{3-11}
&        \tableCoxKeyaGroup            & \multicolumn{1}{c}{{\makecell{0.7308 \\(0.0227)}}} & \multicolumn{1}{c}{\makecell{0.7508 \\(0.0259)}} & \multicolumn{1}{c}{{\makecell{-5.9867 \\(0.0813)}}} & {\makecell{0.0951 \\(0.0024)}} & \multicolumn{1}{c}{\makecell{0.5345 \\(0.1282)}} & \multicolumn{1}{c}{\makecell{0.3053 \\(0.1003)}} & \multicolumn{1}{c}{\makecell{0.8003 \\(0.2040)}} & \multicolumn{1}{c}{\makecell{0.5467 \\(0.1429)}} & \makecell{1.4360 \\(0.3377)}                 \\ \cmidrule{3-11}
&        \tableCoxRPGroup            & \multicolumn{1}{c}{{\makecell{0.7339 \\(0.0196)}}} & \multicolumn{1}{c}{\makecell{0.7548 \\(0.0222)}} & \multicolumn{1}{c}{{\makecell{-5.9960 \\(0.0616)}}} & {\makecell{0.0925 \\(0.0012)}} & \multicolumn{1}{c}{\makecell{0.3985\\ (0.0679)}} & \multicolumn{1}{c}{\makecell{0.1022 \\(0.0835)}} & \multicolumn{1}{c}{\makecell{0.4374 \\(0.0565)}} & \multicolumn{1}{c}{\makecell{0.3127 \\(0.0272)}} & \makecell{1.7580 \\(0.5936)}                 \\ \cmidrule{3-11}
&        \tableCoxKeyaInt           & \multicolumn{1}{c}{{\makecell{0.7280 \\(0.0237)}}} & \multicolumn{1}{c}{\makecell{0.7516 \\(0.0260)}} & \multicolumn{1}{c}{{\makecell{-6.0043 \\(0.0884)}}} & {\makecell{0.0951 \\(0.0016)}} & \multicolumn{1}{c}{\makecell{0.5183 \\(0.0941)}} & \multicolumn{1}{c}{\makecell{0.2781 \\(0.0799)}} & \multicolumn{1}{c}{\makecell{0.6519 \\(0.2615)}} & \multicolumn{1}{c}{\makecell{0.4828 \\(0.1433)}} & \makecell{0.9240 \\(0.4045)}                 \\ \cmidrule{2-11} 
&       \tableDROCox          & \multicolumn{1}{c}{{\makecell{0.7283 \\(0.0054)}}} & \multicolumn{1}{c}{\makecell{0.7494 \\(0.0054)}} & \multicolumn{1}{c}{{\makecell{-6.2140 \\(0.0653)}}} & {\1\textbf{\makecell{0.0880 \\(0.0005)}}} & \multicolumn{1}{c}{\1\textbf{\makecell{0.0791 \\(0.0514)}}} & \multicolumn{1}{c}{\1\textbf{\makecell{0.0113 \\(0.0162)}}} & \multicolumn{1}{c}{\1\textbf{\makecell{0.1317 \\(0.0523)}}} & \multicolumn{1}{c}{\1\textbf{\makecell{0.0740 \\(0.0370)}}} & \1\textbf{\makecell{0.2300 \\(0.2219)}}              \\ \cmidrule{3-11}
&       \tableDROCoxSplit           & \multicolumn{1}{c}{{\makecell{0.7202 \\(0.0137)}}} & \multicolumn{1}{c}{\makecell{0.7404 \\(0.0120)}} & \multicolumn{1}{c}{{\makecell{-6.1406 \\(0.1324)}}} & {\makecell{0.1020 \\(0.0315)}} & \multicolumn{1}{c}{\makecell{0.2150 \\(0.1876)}} & \multicolumn{1}{c}{\1\makecell{0.0392 \\(0.0491)}} & \multicolumn{1}{c}{\1\makecell{0.3115 \\(0.2198)}} & \multicolumn{1}{c}{\1\makecell{0.1885 \\(0.1484)}} & \1\makecell{0.1840 \\(0.1930)}                \\ \cmidrule{1-11} 
\multirow{8}{*}{\rotatebox{90}{Nonlinear \ \ \ \ \ \ \ \ \ \ \ \ \ \ \   }} &          DeepSurv         & \multicolumn{1}{c}{{\textbf{\makecell{0.7488 \\(0.0103)}}}} & \multicolumn{1}{c}{\textbf{\makecell{0.7729 \\(0.0105)}}} & \multicolumn{1}{c}{{\textbf{\makecell{-5.9582 \\(0.0538)}}}} & {\makecell{0.0966 \\(0.0047)}} & \multicolumn{1}{c}{\makecell{0.3686 \\(0.0959)}} & \multicolumn{1}{c}{\makecell{0.1178 \\(0.0771)}} & \multicolumn{1}{c}{\makecell{0.4734 \\(0.1203)}} & \multicolumn{1}{c}{\makecell{0.3199 \\(0.0776)}} & \makecell{0.4270 \\(0.4259)}              \\  \cmidrule{3-11}
&          \tableDeepSurvKeyaInd          & \multicolumn{1}{c}{{\makecell{0.7100 \\(0.0236)}}} & \multicolumn{1}{c}{\makecell{0.7340 \\(0.0237)}} & \multicolumn{1}{c}{{\makecell{-6.1830 \\(0.1620)}}} & {\makecell{0.0995 \\(0.0085)}} & \multicolumn{1}{c}{\textbf{\makecell{0.0564 \\(0.0515)}}} & \multicolumn{1}{c}{\textbf{\makecell{0.0238 \\(0.0198)}}} & \multicolumn{1}{c}{\makecell{1.0618 \\(0.7650)}} & \multicolumn{1}{c}{\makecell{0.3807 \\(0.2370)}} & \makecell{3.0400 \\(1.2074)}               \\  \cmidrule{3-11}
&          \tableDeepSurvRPInd         & \multicolumn{1}{c}{{\makecell{0.7353 \\(0.0104)}}} & \multicolumn{1}{c}{\makecell{0.7579\\(0.0103)}} & \multicolumn{1}{c}{{\makecell{-6.0247 \\(0.0938)}}} & {\makecell{0.0944 \\(0.0057)}} & \multicolumn{1}{c}{\makecell{0.1146 \\(0.0559)}} & \multicolumn{1}{c}{\makecell{0.0297 \\(0.0250)}} & \multicolumn{1}{c}{\makecell{0.6595 \\(0.4998)}} & \multicolumn{1}{c}{\makecell{0.2679 \\(0.1465)}} & \makecell{0.5220 \\(0.2891)}               \\  \cmidrule{3-11}
&        \tableDeepSurvKeyaGroup           & \multicolumn{1}{c}{{\makecell{0.7299 \\(0.0224)}}} & \multicolumn{1}{c}{\makecell{0.7540 \\(0.0210)}} & \multicolumn{1}{c}{{\makecell{-6.0622 \\(0.1643)}}} & {\makecell{0.0972 \\(0.0081)}} & \multicolumn{1}{c}{\makecell{0.2667 \\(0.1408)}} & \multicolumn{1}{c}{\makecell{0.0898 \\(0.0325)}} & \multicolumn{1}{c}{\makecell{0.6532 \\(0.4237)}} & \multicolumn{1}{c}{\makecell{0.3366 \\(0.1456)}} & \makecell{0.7070 \\(0.8405)}             \\ \cmidrule{3-11}
&        \tableDeepSurvRPGroup            & \multicolumn{1}{c}{{\makecell{0.7368 \\(0.0114)}}} & \multicolumn{1}{c}{\makecell{0.7594 \\(0.0109)}} & \multicolumn{1}{c}{{\makecell{-6.0146 \\(0.0916)}}} & {\makecell{0.0942 \\(0.0052)}} & \multicolumn{1}{c}{\makecell{0.1283 \\(0.0593)}} & \multicolumn{1}{c}{\makecell{0.0324 \\(0.0281)}} & \multicolumn{1}{c}{\makecell{0.6080 \\(0.3616)}} & \multicolumn{1}{c}{\makecell{0.2562 \\(0.0999)}} & \makecell{0.5600 \\(0.4160)}              \\ \cmidrule{3-11}
&          \tableDeepSurvKeyaInt         & \multicolumn{1}{c}{{\makecell{0.7344 \\(0.0112)}}} & \multicolumn{1}{c}{\makecell{0.7613 \\(0.0098)}} & \multicolumn{1}{c}{{\makecell{-6.0001 \\(0.0791)}}} & {\makecell{0.0958 \\(0.0047)}} & \multicolumn{1}{c}{\makecell{0.4034 \\(0.1813)}} & \multicolumn{1}{c}{\makecell{0.1209 \\(0.0566)}} & \multicolumn{1}{c}{\makecell{0.4576 \\(0.1760)}} & \multicolumn{1}{c}{\makecell{0.3273 \\(0.1143)}} & \makecell{0.4920 \\(0.4089)}                \\ \cmidrule{2-11}  
&          \tableDeepDROCox          & \multicolumn{1}{c}{{\makecell{0.7305 \\(0.0216)}}} & \multicolumn{1}{c}{\makecell{0.7521 \\(0.0271)}} & \multicolumn{1}{c}{{\makecell{-6.0667 \\(0.1196)}}} & {\1\textbf{\makecell{0.0913 \\(0.0041)}}} & \multicolumn{1}{c}{\makecell{0.2206 \\(0.1239)}} & \multicolumn{1}{c}{\makecell{0.0498 \\(0.0556)}} & \multicolumn{1}{c}{\1\textbf{\makecell{0.3004 \\(0.1191)}}} & \multicolumn{1}{c}{\1\textbf{\makecell{0.1903 \\(0.0932)}}} & \1\textbf{\makecell{0.0910 \\(0.0461)}}            \\ \cmidrule{3-11}
&       \tableDeepDROCoxSplit             & \multicolumn{1}{c}{{\makecell{0.6980 \\(0.0024)}}} & \multicolumn{1}{c}{\makecell{0.7264 \\(0.0027}} & \multicolumn{1}{c}{{\makecell{-6.1182 \\(0.0221)}}} & {\makecell{0.1023 \\(0.0003)}} & \multicolumn{1}{c}{\makecell{0.3781 \\(0.0565)}} & \multicolumn{1}{c}{\makecell{0.1945 \\(0.0386)}} & \multicolumn{1}{c}{\makecell{0.4135 \\(0.0405)}} & \multicolumn{1}{c}{\makecell{0.3287 \\(0.0419)}} & \1\makecell{0.3480 \\(0.2794)}             \\ \bottomrule
\end{tabular}
}
\label{tab:general_performance_SEER_race_CI}
\end{table*}

\begin{figure*}[t!]
\centering
\vspace{-.1em}
\subfigure[\small FLC (age)]{\includegraphics[width=0.32\linewidth,trim={0 1pt 0 16pt},clip]{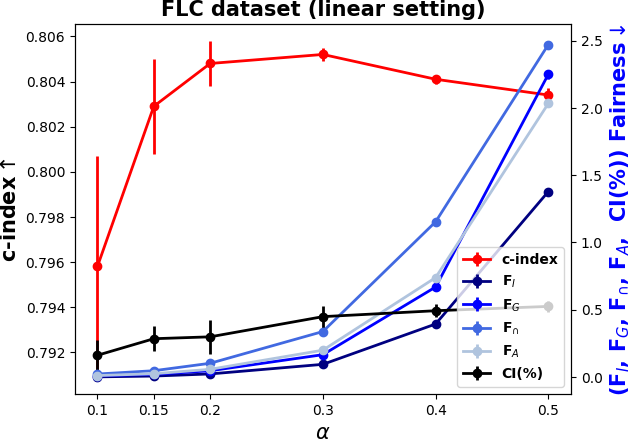}}
\subfigure[\small SUPPORT (age)]{\includegraphics[width=0.325\linewidth,trim={0 1pt 0 16pt},clip]{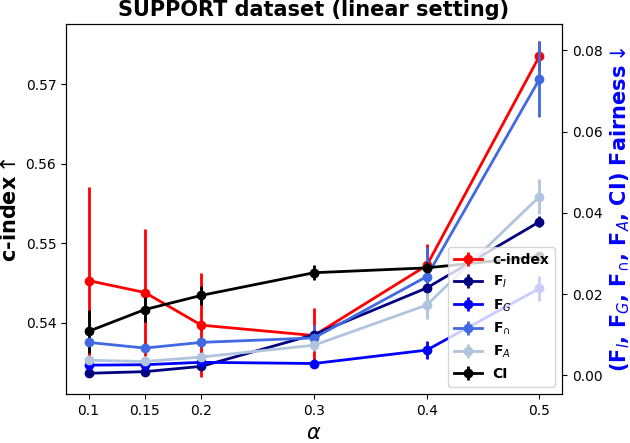}}
\subfigure[\small SEER (age)]{\includegraphics[width=0.32\linewidth,trim={0 1pt 0 16pt},clip]{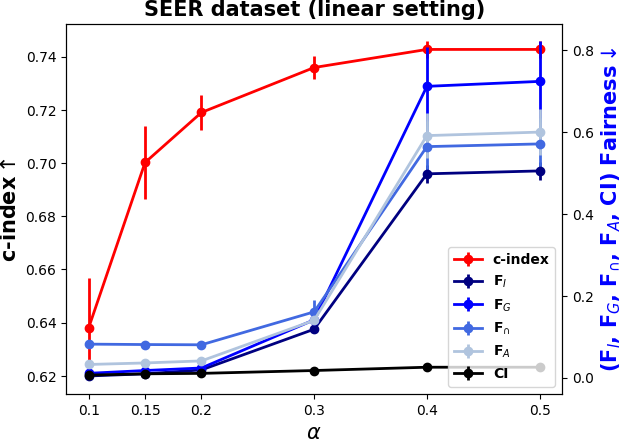}}
\vspace{-.8em}
\caption{\small Effect of $\alpha$ on test set accuracy (c-index; higher is better) and fairness metrics (F$_I$, F$_G$, F$_{\cap}$, F$_A$, and CI; lower is better for all fairness metrics) of \textsc{dro-cox} on three datasets.}
\vspace{-2em}
\label{fig:sensitive_analysis}
\end{figure*}

\smallskip
\noindent
\textbf{Experimental results}. We report the test set evaluation metrics for FLC (using age to evaluate F$_G$ and CI) in Table~\ref{tab:general_performance_CI}, SUPPORT (gender) in Table~\ref{tab:general_performance_SUPPORT_gender_CI}, and SEER (race) in Table~\ref{tab:general_performance_SEER_race_CI}. Experimental results using other sensitive attributes for the datasets have similar trends and are in Appendix~\ref{additional_exp_results}. From these tables, we have the following observations:
\begin{itemize}[leftmargin=*,itemsep=0pt,parsep=0pt,topsep=0pt,partopsep=0pt]
    \item Among linear methods, \textsc{dro-cox} consistently outperforms baselines in terms of the CI fairness metric (and often on the other fairness metrics too) while still achieving reasonably high accuracy scores. A similar trend holds among nonlinear methods for the deep \textsc{dro-cox} variant.

    \item The performance difference (in terms of both accuracy and fairness) between \textsc{dro-cox} and \textsc{dro-cox (split)} is not clear cut; sometimes one performs better than the other and vice versa. This holds for their linear variants as well as, separately, their nonlinear (deep) variants.
    
    \item As expected, the unregularized Cox and DeepSurv models often have (among) the highest accuracy scores but tend to have poor performance on fairness metrics. %
    
    \item The baselines that are regularized variants of Cox and DeepSurv typically do not simultaneously achieve low scores across all fairness metrics. Even though some of these can work well with some of the metrics by \citet{keya2021equitable}, they clearly do not work as well as our \textsc{dro-cox} variants when it comes to the CI fairness metric that actually accounts for accuracy.
\end{itemize}
\textbf{Effect of $\alpha$}.
To show how $\alpha$ trades off between fairness and accuracy, we show results for \mbox{\textsc{dro-cox}} in the linear setting across all datasets (using age for evaluating F$_G$ and CI) in Figure~\ref{fig:sensitive_analysis}, where we use c-index as the accuracy metric. It is clear that accuracy tends to increase when $\alpha$ increases from 0.1 to 0.3 on FLC and SEER, and from 0.3 to 0.5 on SUPPORT. However, the increase in $\alpha$ results in worse scores across fairness metrics.

\noindent
\textbf{Additional experiments.} Across all methods, instead of minimizing the validation set CI fairness metric during hyperparameter tuning (tolerating a small degradation in validation set c-index), we also tried instead minimizing the validation set F$_A$ metric and found similar results: our \textsc{dro-cox} variants end up consistently outperforming all the baselines on F$_A$ (and also often achieves competitive CI metric scores) without a large accuracy drop. We also show that our \textsc{dro-cox (split)} procedure is somewhat robust to the choice of $n_1$ and $n_2$, and if \textsc{dro-cox (split)} did not use both losses $\mathcal{L}_{\text{DRO-split}}(\theta, \eta, \mathcal{D}_1, \mathcal{D}_2)$ and $\mathcal{L}_{\text{DRO-split}}(\theta, \eta', \mathcal{D}_2, \mathcal{D}_1)$ (i.e., if it only used one of these), then it performs worse. For details on these experiments, see Appendix~\ref{additional_exp_results}.

\section{Discussion}
\vspace{-.4em}

We have shown how to apply DRO to Cox models in a manner that is compliant with existing DRO theory (\textsc{dro-cox (split)}) and in a manner that is heuristic (\textsc{dro-cox}). Importantly, how we applied DRO to Cox models works with other survival models as well. The key idea is to write the overall loss in terms of individual losses, which in turn could be used in a DRO framework. An open question is whether we could derive a theoretically sound \mbox{\textsc{dro-cox}} variant that does not require sample splitting. This same technical challenge would arise in working with other survival models that use pairwise comparisons between patients. When a parametric survival model is used in which each patient's loss does not depend on other patients, we point out that existing DRO machinery directly works; a strategy such as sample splitting would be unnecessary. We defer a thorough evaluation of DRO applied to more survival models to future work.

\vspace{-1.1em}
\section*{Acknowledgments}
\vspace{-.4em}

This work was supported by NSF CAREER award \#2047981. The authors would like to thank Tatsunori Hashimoto and the anonymous reviewers for very helpful feedback.

\bibliography{hu22}

\appendix
\numberwithin{equation}{section}
\numberwithin{theorem}{section}
\numberwithin{figure}{section}
\numberwithin{table}{section}
\renewcommand{\thesection}{{\Alph{section}}}
\renewcommand{\thesubsection}{\Alph{section}.\arabic{subsection}}
\renewcommand{\thesubsubsection}{\Roman{section}.\arabic{subsection}.\arabic{subsubsection}}

\def\p{\mathbf{p}}
\def\v{\mathbf{v}}
\def\u{\mathbf{u}}

\vspace{-.75em}
\section{Estimating the Baseline Hazard and Survival Function}
\label{sec:breslow}
\vspace{-.25em}

After learning the log partial hazard function $f(\cdot;\theta)$ (or, equivalently, learning the parameters $\theta$), a standard approach to estimating the baseline hazard function $h_0$ is to use the so-called Breslow method \citep{breslow1972discussion}. In what follows, we use $\widehat{\theta}$ to denote the learned estimate of $\theta$.

The Breslow method estimates a discretized version of $h_0$. Specifically, let $t_1 < t_2 < \cdots < t_m$ denote the unique times when critical event happened in the training data. Let $d_j$ denote the number of critical events that occurred at time $t_j$, where $j=1,\dots,m$. Then we compute the following estimate of $h_0$ at the $j$-th time step:
\[
\widehat{h}_{0,j}
:= \frac{d_j}{\sum_{i=1}^n \mathbf{1}\{Y_i \ge t_j\}\exp(f(x_i;\widehat{\theta}))}.
\]
After estimating the baseline hazard function, estimating the survival function is straightforward. Recall that $S(t|x)=\exp\Big(-\int_0^t h(u|x)du\Big)$. Then combining this equation with the factorization~\eqref{eq:hazard-factorization}, we get
\begin{align}
S(t|x)
&= \exp\Big(-\int_0^t h_0(u)\exp(f(x;\theta)) du \Big) \nonumber \\
&= \exp\Big(\Big[-\!\!\!\!\!\!\underbrace{\int_0^t h_0(u)du}_{\text{abbreviate as }H_0(t)}\!\!\!\!\!\!\Big] \exp(f(x;\theta)) \Big).
\label{eq:how-to-estimate-S-helper}
\end{align}
We can estimate $H_0(t)$ via a summation in place of an integration:
\[
\widehat{H}_0(t):=\sum_{j=1}^m \mathbf{1}\{t_j \le t\}\widehat{h}_{0,j}.
\]
Thus, by plugging in $\widehat{H}_0$ in place of $H_0$ and $\widehat{\theta}$ in place of $\theta$ in equation~\eqref{eq:how-to-estimate-S-helper}, we obtain the survival function estimate $\widehat{S}(t|x):=\exp(-\widehat{H}_0(t)\exp(f(x;\widehat{\theta})))$.

\vspace{-.75em}
\section{Fairness Metrics}
\label{sec:fairness-measures}
\vspace{-.25em}

In this paper, we use the individual, group, and intersectional fairness metrics defined by \citet{keya2021equitable} and also the concordance imparity (CI) metric by \citet{zhang2022longitudinal}. In what follows, since we are focusing on Cox proportional hazards models, we can take the predicted outcome for a feature vector $x$ to be the so-called \emph{partial hazard} $\widetilde{h}(x):=\exp( f(x;\theta) )$; this is the same as the hazard function given in equation \eqref{eq:hazard-factorization} except where we exclude the baseline hazard factor $h_0(t)$. Note that once we exclude $h_0(t)$, then $\widetilde{h}$ no longer depends on time~$t$. We state the fairness metrics in terms of a collection of $N_{\text{test}}$ test patients with data $(X_1^{\text{test}},Y_1^{\text{test}},\delta_1^{\text{test}}),\dots,(X_{N_{\text{test}}}^{\text{test}},Y_{N_{\text{test}}}^{\text{test}},\delta_{N_{\text{test}}}^{\text{test}})$. Note that the fairness metrics by \citet{keya2021equitable} only use the test feature vectors $X_1^{\text{test}},\dots,X_{N_{\text{test}}}^{\text{test}}$ and ignores the test patients' observed times and event~indicators. Also, at the end of this section, we point out that the individual and group fairness metrics by \citet{keya2021equitable} are sensitive to the scale of the log partial hazard~$f(\cdot;\theta)$.

\smallskip
\noindent
\textbf{Individual fairness.}
Roughly, \citet{keya2021equitable} consider a model to be fair across individuals (patients) if similar individuals have similar predicted outcomes. To operationalize this notion of fairness in the context of Cox models, Keya et al.~define the individual fairness metric
\begin{equation*}
\begin{aligned}
\textrm{F}_I:=\sum_{i=1}^{N_{\text{test}}}\sum_{j=i+1}^{N_{\text{test}}}\big[&|\widetilde{h}(X_i^{\text{test}})-\widetilde{h}(X_j^{\text{test}})|\\[-.75em]
&-\gamma \|X_i^{\text{test}} - X_j^{\text{test}}\|\big]_+,
\end{aligned}
\end{equation*}
where $\gamma$ is a predefined scale factor (0.01 in our experiments). As a reminder, $[\cdot]_+$ is the ReLU function (so that $[a]_+=\max\{0,a\}$ for any $a\in\mathbb{R}$).

Note that this individual fairness metric is actually just penalizing $\widetilde{h}$ for not being Lipschitz continuous (as empirically evaluated over the test data). Specifically, $\widetilde{h}$ is defined to be $\gamma$-Lipschitz continuous if
\[
|\widetilde{h}(x)-\widetilde{h}(x')|
\le \gamma \|x - x'\|
\quad\text{for all }x,x'\in\mathcal{X}.
\]
Meanwhile, when F$_I$ is equal to 0, then it means that
\begin{align*}
&|\widetilde{h}(X_i^{\text{test}})-\widetilde{h}(X_j^{\text{test}})|
\le \gamma \|X_i^{\text{test}} - X_j^{\text{test}}\| \\
&\qquad\text{for all }i,j\in\{1,\dots,N_{\text{test}}\}.
\end{align*}
As a technical remark, in the definition of F$_I$ and also $\gamma$-Lipschitz continuity, the metric used to measure distances between feature vectors does not have to be Euclidean. For example, we can replace $\|X_i^{\text{test}} - X_j^{\text{test}}\|$ with $\rho(X_i^{\text{test}}, X_j^{\text{test}})$, where $\rho:\mathcal{X}\times\mathcal{X}\rightarrow[0,\infty)$ is a user-specified metric.

\smallskip
\noindent
\textbf{Group fairness.}
Next, \citet{keya2021equitable} consider a model is fair across a user-specified set of groups if these different groups have similar predicted outcomes. Keya et al.~define the group fairness metric F$_G$ to look at the maximum deviation of a group's average predicted outcome to the overall population's average predicted outcome. Specifically, let $\mathcal{G}$ be the user-specified set of groups to consider (for example, there could be two groups: everyone with age at most 65 years, and everyone older than 65 years), where each group $g\in\mathcal{G}$ is a subset of the test set indices $\{1,\dots,N_{\text{test}}\}$ (so that using this notation, group $g$ has size $|g|$); the different groups should form a partition of the test set (so that the groups are disjoint and their union is the entire test set). Then
\begin{equation*}
\begin{aligned}
&\textrm{F}_G
:= \\
&\max_{g\in \mathcal{G}}
    \bigg|
      \underbrace{\frac{1}{|g|}\sum_{i\in g}\widetilde{h}(X_i^{\text{test}})}_{\substack{\text{average predicted}\\\text{outcome of group }g}}
      -
      \underbrace{\frac{1}{N_{\text{test}}}\sum_{i=1}^{N_{\text{test}}}\widetilde{h}( X_i^{\text{test}})}_{\substack{\text{average predicted}\\\text{outcome of population}}}
    \bigg|.
\end{aligned}
\end{equation*}

\smallskip
\noindent
\textbf{Intersectional fairness.}
\citet{keya2021equitable} consider a notion of intersectional fairness that accounts for multiple sensitive attributes. For example, in the FLC dataset, we have 2 different sensitive attributes, age and gender. For each of these sensitive attributes, we can partition the test set into groups. Specifically, let $\mathcal{G}_1$ be a partition of the test set into different age groups (for example, two groups: at most 65 years old and over 65 years old), and let $\mathcal{G}_2$ be a partition of the test set into different gender groups (for example, two groups: female and male). Then intersectional fairness looks at every intersection of age/gender groups (continuing from the previous examples, we would have four intersectional subgroups: at most 65 years old and female, at most 65 years and male, over 65 years old and female, over 65 years old and male).

The notation here is a bit more involved. The set of all intersectional subgroups of $\mathcal{G}_1$ and $\mathcal{G}_2$ is given by the Cartesian product $\mathcal{G}_1\times\mathcal{G}_2$. Note that $s\in\mathcal{G}_1\times\mathcal{G}_2$ means that $s=(s_1,s_2)$, where $s_1\in\mathcal{G}_1$ and $s_2\in\mathcal{G}_2$. More generally, if there are $J$ sensitive attributes, corresponding to groupings $\mathcal{G}_1,\mathcal{G}_2,\dots,\mathcal{G}_J$, then the set of all intersectional subgroups would be $\mathcal{S}:=\mathcal{G}_1\times\mathcal{G}_2\times\cdots\mathcal{G}_J$. Now $s\in\mathcal{S}$ is a list consisting of $J$ different subsets of test patients (i.e., $s=(s_1,s_2,\dots,s_J)$, where $s_1\in\mathcal{G}_1$, $\dots$, $s_J\in\mathcal{G}_J$). The intersection of these $J$ subsets (i.e., $\cap_{j=1}^J s_j \subset \{1,\dots,N_{\text{test}}\}$) is precisely the set of test patients that intersectional subgroup $s$ corresponds to. Then the average predicted outcome for intersectional subgroup $s$ is
\[
\widetilde{\mathbf{h}}(s):=
\frac{1}{|\cap_{j=1}^J s_j|}~ \sum_{i\in\cap_{j=1}^J s_j} \widetilde{h}(X_i^{\text{test}}).
\]
Then the intersection fairness metric F$_{\cap}$ by
\citet{keya2021equitable} is the worst-case log ratio of expected predicted outcomes between two intersectional subgroups:
\begin{equation*} %
\textrm{F}_{\cap}:=\max_{s, s' \in \mathcal{S}}
\Big|
\log \frac{ \widetilde{\mathbf{h}}(s) }
          { \widetilde{\mathbf{h}}(s') }
\Big|.
\end{equation*}
\noindent
\textbf{Concordance imparity}. We now describe an alternative metric for group fairness called concordance imparity (CI) that asks that a survival analysis model achieves similar prediction accuracy for different groups. For ease of exposition, we only state the CI metric by \citet{zhang2022longitudinal} in terms of a single sensitive attribute that has already been discretized (e.g., the attribute is already discrete or we have a pre-specified discretization rule); this special case is sufficient for our experiments. We denote the set of possible discretized values of this sensitive attribute as $\mathcal{A}$. For example, $\mathcal{A}$ could correspond to age and we could have $\mathcal{A}=\{\text{``age$\le$65''},\text{``age$>65$''}\}$, i.e., $\mathcal{A}$ consists of the different groups to consider. We refer the reader to the Zhang and Weiss's original paper for their more general definition of CI that can handle a continuous sensitive attribute via an automatic discretization strategy that they propose.

Assuming that the sensitive attribute has already been discretized into the set $\mathcal{A}$, the CI metric looks at a variant of the standard survival analysis accuracy metric of concordance index \citep{harrell1982evaluating} that Zhang and Weiss call the \emph{concordance fraction} (CF), which is specific to each sensitive attribute value $a\in\mathcal{A}$. The CI metric is then defined to be the worst-case difference between the CF scores of any two $a,a'\in\mathcal{A}$ where $a\ne a'$. The pseudocode can be found in Algorithm \ref{alg:CI-discrete}; note that to keep the notation from getting clunky, we drop the superscript ``test'' from the test feature vectors, observed times, and event indicators in the pseudocode but we still use $N_{\text{test}}$ to denote the number of test patients. Also, in the pseudocode, we let $A_i\in\mathcal{A}$ denote the sensitive attribute value for the $i$-th test patient, where we assume that $A_i$ can directly be computed based on the $i$-th test patient's feature vector. For example, when age (which is not discretized) is one of the features and $\mathcal{A}$ consists of the two age groups previously stated ($\le65$ or $>65$), then since we know the discretization rule used, we can readily determine which age group in $\mathcal{A}$ that any test patient is in.

\begin{algorithm}[p!]\scriptsize
    \caption{Concordance Imparity (CI) with a discrete sensitive attribute\vspace{1pt}}\label{alg:CI-discrete}
    \SetAlgoLined
    \KwIn{Test dataset $\{(X_i,Y_i,\delta_i)\}_{i=1}^{N_{\text{test}}}$, risk score $f(\cdot;\theta)$ (from an already trained model),
    set of sensitive attribute values $\mathcal{A}$ (so that each $a\in\mathcal{A}$ corresponds to a different group), $A_1,\dots,A_{N_{\text{test}}}\in\mathcal{A}$ says which sensitive attribute value each test patient has}
    \KwOut{CI score} 

    \For{$a\in\mathcal{A}$}{
        Initialize the numerator count $\mathbf{N}(a) \leftarrow 0$ and denominator count $\mathbf{D}(a) \leftarrow 0$.
    }
    \For{$i=1,\dots,N_{\text{test}}$}{
    \For{$j=1,\dots,N_{\text{test}}$ s.t.~$j\ne i$}{
        \eIf{($Y_i<Y_j$ and $\delta_i==0$) or ($Y_j<Y_i$ and $\delta_j==0$) or ($Y_i==Y_j$ and $\delta_i==0$ and $\delta_j==0$)}
        {
        \textbf{continue}}{
        Set $\mathbf{D}(A_i) \leftarrow \mathbf{D}(A_i) + 1$.
        }
        \uIf{$Y_i<Y_j$}{
            \uIf{$f(X_i;\theta)>f(X_j;\theta)$}{
            Set $\mathbf{N}(A_i) \leftarrow \mathbf{N}(A_i) + 1$.
            }
            \ElseIf{$f(X_i;\theta)==f(X_j;\theta)$}{
            Set $\mathbf{N}(A_i) \leftarrow \mathbf{N}(A_i) + 0.5$.
            }{}
        }
        \uElseIf{$Y_i>Y_j$}{
            \uIf{$f(X_i;\theta)<f(X_j;\theta)$}{
            Set $\mathbf{N}(A_i) \leftarrow \mathbf{N}(A_i) + 1$.
            }
            \ElseIf{$f(X_i;\theta)==f(X_j;\theta)$}{
            Set $\mathbf{N}(A_i) \leftarrow \mathbf{N}(A_i) + 0.5$.
            }{}
        }
        \ElseIf{$Y_i==Y_j$}{
            \uIf{$\delta_i==1$ and $\delta_j==1$}{\eIf{$f(X_i;\theta)\!\!==\!\!f(X_j;\theta)$}{%
            Set $\mathbf{N}(A_i) \leftarrow \mathbf{N}(A_i) + 1$.
            }{%
            Set $\mathbf{N}(A_i) \leftarrow \mathbf{N}(A_i) + 0.5$.
            }}
            \uElseIf{$\delta_i\!\!==\!\!0$ and $\delta_j\!\!==\!\!1$ and $f(X_i;\theta)\!<\!f(X_j;\theta)$}{%
            Set $\mathbf{N}(A_i) \leftarrow \mathbf{N}(A_i) + 1$.
            }
            \uElseIf{$\delta_i\!\!==\!\!1$ and $\delta_j\!\!==\!\!0$ and $f(X_i;\theta)\!>\!f(X_j;\theta)$}{%
            Set $\mathbf{N}(A_i) \leftarrow \mathbf{N}(A_i) + 1$.
            }
            \Else{%
            Set $\mathbf{N}(A_i) \leftarrow \mathbf{N}(A_i) + 0.5$.
            }
        }{}
    }

    }
    \For{$a\in\mathcal{A}$}{
        Set the concordance fraction of $a$: $\mathbf{CF}(a) \leftarrow \frac{\mathbf{N}(a)}{\mathbf{D}(a)}$.
    }
    
    \Return{$\text{\emph{CI}}\leftarrow\max_{a,a'\in\mathcal{A}\text{ s.t.~}a\ne a'}|\mathbf{CF}(a)-\mathbf{CF}(a')|$}
\end{algorithm}

\vspace{-.75em}
\subsection*{Scale Issues with F$_I$ and F$_G$}
\vspace{-.25em}

We point out that the F$_I$ and F$_G$ fairness metrics are sensitive to the scale of the log partial hazard function $f(\cdot;\theta)$, and thus also the scale of the partial hazard $\widetilde{h}(x)=\exp( f(x;\theta) )$. For instance, consider a standard linear Cox model with $f(x;\theta)=\theta^T x$, where the parameters $\theta$ have already been learned. Then one way to make the model appear fairer according to the F$_I$ and F$_G$ metrics is to just scale all values in $\theta$ by any positive constant smaller than 1; doing so, the standard accuracy metric of concordance index \citep{harrell1982evaluating} would actually remain unchanged for the model as it only depends on the ranking of the different individuals' (log) partial hazard values. However, an accuracy score that considers each individual's survival function estimate (e.g., integrated IPCW Brier Score \citep{graf1999assessment}) would be affected.

\vspace{-.75em}
\section{Pseudocode for Our Proposed Methods}
\label{sec:pseudocode}
\vspace{-.25em}

We provide pseudocode for \textsc{dro-cox} and \textsc{dro-cox (split)} in Algorithm \ref{alg:dro-cox} and Algorithm \ref{alg:dro-cox-split}, respectively.

\begin{algorithm}[p!]\scriptsize
    \caption{\textsc{dro-cox}}\label{alg:dro-cox}
    \SetAlgoLined
    \KwIn{A training dataset $\{(X_i,Y_i,\delta_i)\}_{i=1}^n$, minimum subpopulation probability hyperparameter $\alpha$, learning rate $\xi$, max\_iterations}
    \KwOut{Survival model parameters $\widehat{\theta}$} 
    Obtain initial survival model parameters $\widehat{\theta}_0$ (e.g., using default PyTorch parameter initialization).
    
    \For{$l=0$ to \emph{max\_iterations}}{
    \For{$i=1$ to $n$}{
        Set $u_i \leftarrow \ell_i(\widehat{\theta}_l)$ using equation~\eqref{eq:cox-individual-loss}.
    }
    Set $\widehat{\eta}$ to be the value of $\eta\in\mathbb{R}$ that minimizes $\mathcal{L}_{\text{DRO}}(\widehat{\theta}_l, \eta)$ as given in equation~\eqref{eq:empirical_dro}, where in the empirical average, the $i$-th individual's loss is set to be the variable $u_i$ computed above. This minimization is solved using binary search.
    
    Set $\widehat{\theta}_{l+1} \leftarrow \widehat{\theta}_l -\xi\cdot\nabla_{\theta}{\mathcal{L}}_{DRO}(\widehat{\theta}_l, \widehat{\eta})$.
    
    }
    \Return{$\widehat{\theta} \leftarrow \widehat{\theta}_{\emph{max\_iterations}+1}$}
\end{algorithm}

\begin{algorithm}[p!]\scriptsize
    \caption{\textsc{dro-cox (split)}}\label{alg:dro-cox-split}
    \SetAlgoLined
    \KwIn{A training dataset $\{(X_i,Y_i,\delta_i)\}_{i=1}^n$, minimum subpopulation probability hyperparameter $\alpha$, $n_1$, learning rate $\xi$, max\_iterations}
    \KwOut{Survival model parameters $\widehat{\theta}$} 
    Obtain initial survival model parameters $\widehat{\theta}_0$ (e.g., using default PyTorch parameter initialization).
    
    Set $\mathcal{D}_1 \leftarrow \{1,2,\dots,n_1\}$ and $\mathcal{D}_2 \leftarrow \{n_1+1,\dots,n\}$.
    
    \For{$l=0$ to \emph{max\_iterations}}{
    \For{$i\in \mathcal{D}_1$}{
        Set $u_i \leftarrow \widetilde{\ell}_{\text{split}}(\widehat{\theta}_l;X_i,Y_i,\delta_i,\mathcal{D}_2)$ with equation~\eqref{eq:split_individual_loss}.
    }
    Set $\widehat{\eta}$ to be the value of $\eta\in\mathbb{R}$ that minimizes $\mathcal{L}_{\text{DRO-split}}(\widehat{\theta}_l, \eta, \mathcal{D}_1, \mathcal{D}_2)$ as given in equation~\eqref{eq:empirical_dro_split}, where in the empirical average, the $i$-th individual's loss is set to be the variable $u_i$ computed above. This minimization is solved using binary search.
    
    \For{$i\in\mathcal{D}_2$}{
        Set $v_i \leftarrow \widetilde{\ell}_{\text{split}}(\widehat{\theta}_l;X_i,Y_i,\delta_i,\mathcal{D}_1)$ with equation~\eqref{eq:split_individual_loss}.
    }

    Set $\widehat{\eta}'$ to be the value of $\eta'\in\mathbb{R}$ that minimizes $\mathcal{L}_{\text{DRO-split}}(\widehat{\theta}_l, \eta', \mathcal{D}_2, \mathcal{D}_1)$ as given in equation~\eqref{eq:empirical_dro_split}, where in the empirical average, the $i$-th individual's loss is set to be the variable $v_i$ computed above. This minimization is solved using binary search.
    
    Set $\widehat{\theta}_{l+1} \leftarrow \widehat{\theta}_l -\xi \cdot\big(\nabla_{\theta}{\mathcal{L}}_{\text{DRO-split}}(\widehat{\theta}_l, \widehat{\eta},\mathcal{D}_1,\mathcal{D}_2)+\nabla_{\theta}{\mathcal{L}}_{\text{DRO-split}}(\widehat{\theta}_l, \widehat{\eta}',\mathcal{D}_2,\mathcal{D}_1)\big)$.
    
    }
    \Return{$\widehat{\theta} \leftarrow \widehat{\theta}_{\emph{max\_iterations}+1}$}
\end{algorithm}

\vspace{-.75em}
\section{Hyperparameter Tuning and Compute Environment Details}
\label{sec:hyperparameters-compute-env}
\vspace{-.25em}

\textbf{Hyperparameters.}
For nonlinear Cox models, we always use a two-layer MLP with ReLU as the activation function and 24 as the number of hidden units. All models (linear and nonlinear) are trained using Adam \citep{kingma2014adam} in PyTorch 1.7.1 in a batch setting for 500 iterations, only using a CPU and no GPU. We tune on the following hyperparameter grid:
\begin{itemize}[leftmargin=*,itemsep=0pt,parsep=0pt,topsep=0pt,partopsep=0pt]
\item learning rate: 0.01, 0.001, 0.0001
\item $\lambda$ (only used for baselines; a hyperparameter that controls the tradeoff between the original Cox loss and fairness regularization term): 1, 0.7, 0.4
\item $\alpha$ (for \textsc{dro-cox}/\textsc{dro-cox (split)} variants): 0.1, 0.15, 0.2, 0.3, 0.4, 0.5
\end{itemize}
In addition, for \textsc{dro-cox (split)}, we choose $n_1=n_2=n/2$ (rounding as needed when $n$ is odd, so that $n_1$ might not equal $n_2$).

Following \citet{keya2021equitable}, the final hyperparameter setting per dataset and per method is determined based on a preset rule that allows up to a 5\% degradation in the validation set c-index from the classical Cox model (for the linear setting) or DeepSurv (for the nonlinear setting) while minimizing the validation set CI fairness metric.

\smallskip
\noindent
\textbf{Compute environment.}
All models are implemented with Python 3.8.3, and they are trained and tested on identical compute instances, each with an Intel Core i9-10900K CPU (3.70GHz with 64 GB RAM). As a reminder, we did not train using a GPU.

\vspace{-.75em}
\section{Additional Experiments}\label{additional_exp_results}
\vspace{-.25em}

\textbf{Using other sensitive attributes in evaluating F$_G$ and CI.} In the main paper, we only showed test set performance metrics for FLC, SUPPORT, and SEER using age, gender, and race respectively in evaluating F$_G$ and CI. We now provide results using gender for FLC (Table~\ref{tab:general_performance_FLC_gender_CI}), age and separately race for SUPPORT (Tables~\ref{tab:general_performance_SUPPORT_age_CI} and~\ref{tab:general_performance_SUPPORT_race_CI}), and age for SEER (Table~\ref{tab:general_performance_SEER_age_CI}). Our main findings still hold for these additional results.

\smallskip
\noindent
\textbf{Hyperparameter tuning based on F$_A$ instead of CI.}
The previous experimental results are based on hyperparameters chosen by minimizing the validation set CI fairness metric (while tolerating a small degradation in c-index). If instead of focusing on the CI fairness metric, we used F$_A$ instead, then we get the results in Tables \ref{tab:general_performance}, \ref{tab:general_performance_FLC_gender}, \ref{tab:general_performance_SUPPORT_age}, \ref{tab:general_performance_SUPPORT_race},
\ref{tab:general_performance_SUPPORT_gender}, \ref{tab:general_performance_SEER_age}, 
and \ref{tab:general_performance_SEER_race}. In particular, our experimental findings from before remain the same except now our \textsc{dro-cox} and \textsc{dro-cox (split)} variants consistently achieve the best F$_A$ scores (while often also scoring well on other fairness metrics) without too large of a drop in accuracy.

\smallskip
\noindent
\textbf{Effect of changing $n_1$ (or $n_2$) for }\mbox{\textsc{dro-cox (split)}}\textbf{.} In the above experiments, we set $n_1=n_2=n/2$ (rounding as needed). To evaluate the sensitivity of this setting, we test the model performance using \textsc{dro-cox (split)} under the linear and nonlinear settings, where we set $n_2=0.1n,0.2n,0.3n,0.4n,0.5n$ (corresponding to $n_1=0.9n,0.8n,0.7n,0.6n,0.5n$). We report the test set performance metrics for the FLC dataset (using age in evaluating F$_G$ and CI) in Table \ref{tab:performance_of_diff_split}. From the table, we find that per metric, different settings for $n_1$ and $n_2$ lead to results that, while slightly different, are not dramatically different, i.e., the performance of \textsc{dro-cox (split)} does not appear very sensitive w.r.t.~the choice of $n_1$ and $n_2$.

\smallskip
\noindent
\textbf{The effect of using two losses for }\mbox{\textsc{dro-cox (split)}}\textbf{ rather than only one.} Recall that \textsc{dro-cox (split)} minimizes the sum of two losses $\mathcal{L}_{\text{DRO-split}}(\theta, \eta, \mathcal{D}_1, \mathcal{D}_2)$ and $\mathcal{L}_{\text{DRO-split}}(\theta, \eta', \mathcal{D}_2, \mathcal{D}_1)$. Towards the end of Section~\ref{sec:individual-loss}, we said that an approach that only minimizes one of these losses would not use the data as effectively compared to minimizing the sum of these losses. We conducted an experiment to verify this claim, where we refer to the version of \textsc{dro-cox (split)} that only minimizes $\mathcal{L}_{\text{DRO-split}}(\theta, \eta, \mathcal{D}_1, \mathcal{D}_2)$ as \mbox{\textsc{dro-cox (split, one side)}}. Specifically, we compare \mbox{\textsc{dro-cox (split, one side)}} and \textsc{dro-cox (split)} under linear and nonlinear settings on the FLC dataset using age to evaluate F$_G$ and CI. We report the resulting test set performance metrics in Table~\ref{tab:performance_of_one_side_split}. From the table, we find that \textsc{dro-cox (split)} outperforms \textsc{dro-cox (split, one side)} on most metrics. This experimental finding supports our hypothesis that \textsc{dro-cox (split, one side)} uses data less effectively.

\onecolumn

\begin{table}[H]
\caption{\small Test set scores on the FLC (gender) dataset, in the same format as Table~\ref{tab:general_performance_CI}.} 
\centering
\setlength\tabcolsep{3.6pt}
{\tiny %
\renewcommand{\belowrulesep}{0.1pt}
\renewcommand{\aboverulesep}{0.1pt}
\begin{tabular}{cccccc|ccccc}
\toprule
\multirow{2}{*}{} & \multirow{2}{*}{Methods} & \multicolumn{4}{c|}{Accuracy Metrics}                                                    & \multicolumn{5}{c}{Fairness Metrics}                             \\ \cmidrule{3-11}
&                   & \multicolumn{1}{c}{c-index$\uparrow$} & \multicolumn{1}{c}{AUC$\uparrow$} & \multicolumn{1}{c}{LPL$\uparrow$} & IBS$\downarrow$ & \multicolumn{1}{c}{F$_I$$\downarrow$} & \multicolumn{1}{c}{F$_G$$\downarrow$} & \multicolumn{1}{c}{F$_{\cap}$$\downarrow$}& \multicolumn{1}{c}{F$_A$$\downarrow$} &CI(\%)$\downarrow$                  \\ \midrule
\multirow{8}{*}{\rotatebox{90}{Linear \ \ \ \ \ \ \ \ \ \ \ \ \ \ }} &        \tableCox            & \multicolumn{1}{c}{{\textbf{\makecell{0.8032 \\(0.0002)}}}} & \multicolumn{1}{c}{\makecell{0.8176 \\(0.0005)}} & \multicolumn{1}{c}{{\textbf{\makecell{-6.3724 \\(0.0011)}}}} & {\makecell{0.1739 \\(0.0004)}} & \multicolumn{1}{c}{\makecell{1.8787 \\(0.0304)}} & \multicolumn{1}{c}{\makecell{0.5421 \\(0.0299)}} & \multicolumn{1}{c}{\makecell{2.8355 \\(0.0297)}} & \multicolumn{1}{c}{\makecell{1.7521 \\(0.0266)}} & \makecell{0.8610 \\(0.0197)}               \\  \cmidrule{3-11}
&          \tableCoxKeyaInd         & \multicolumn{1}{c}{{\makecell{0.7932 \\(0.0083)}}} & \multicolumn{1}{c}{\makecell{0.8175 \\(0.0078)}} & \multicolumn{1}{c}{{\makecell{-6.6845 \\(0.0786)}}} & {\makecell{0.1368 \\(0.0052)}} & \multicolumn{1}{c}{\makecell{0.3234 \\(0.1098)}} & \multicolumn{1}{c}{\makecell{0.0273 \\(0.0214)}} & \multicolumn{1}{c}{\makecell{0.9325 \\(0.2562)}} & \multicolumn{1}{c}{\makecell{0.4277 \\(0.1249)}} & \makecell{1.6750 \\(0.7969)}                \\
\bottomrule
\end{tabular}
}
\label{tab:general_performance_FLC_gender_CI}
\end{table}

\begin{table}[H]
\caption{\small Test set scores on the SUPPORT (age) dataset, in the same format as Table~\ref{tab:general_performance_CI}.}
\centering
\setlength\tabcolsep{3.6pt}
{\tiny %
\renewcommand{\belowrulesep}{0.1pt}
\renewcommand{\aboverulesep}{0.1pt}
\begin{tabular}{cccccc|ccccc}
\toprule
\multirow{2}{*}{} & \multirow{2}{*}{Methods} & \multicolumn{4}{c|}{Accuracy Metrics}                                                    & \multicolumn{5}{c}{Fairness Metrics}                             \\ \cmidrule{3-11}
&                   & \multicolumn{1}{c}{c-index$\uparrow$} & \multicolumn{1}{c}{AUC$\uparrow$} & \multicolumn{1}{c}{LPL$\uparrow$} & IBS$\downarrow$ & \multicolumn{1}{c}{F$_I$$\downarrow$} & \multicolumn{1}{c}{F$_G$$\downarrow$} & \multicolumn{1}{c}{F$_{\cap}$$\downarrow$}& \multicolumn{1}{c}{F$_A$$\downarrow$} &CI(\%)$\downarrow$                  \\ \midrule
\multirow{8}{*}{\rotatebox{90}{Linear \ \ \ \ \ \ \ \ \ \ \ \ \ \ }} &        \tableCox            & \multicolumn{1}{c}{{\textbf{\makecell{0.6025 \\(0.0005)}}}} & \multicolumn{1}{c}{\makecell{0.6163 \\(0.0010)}} & \multicolumn{1}{c}{{\textbf{\makecell{-6.8761 \\(0.0010)}}}} & {\makecell{0.2304 \\(0.0015)}} & \multicolumn{1}{c}{\makecell{0.2113 \\(0.0093)}} & \multicolumn{1}{c}{\makecell{0.1528 \\(0.0059)}} & \multicolumn{1}{c}{\makecell{0.4490 \\(0.0322)}} & \multicolumn{1}{c}{\makecell{0.2710 \\(0.0128)}} & \makecell{2.2240 \\(0.1078)}              \\  \cmidrule{3-11}
&          \tableCoxKeyaInd          & \multicolumn{1}{c}{{\makecell{0.5820 \\(0.0116)}}} & \multicolumn{1}{c}{\makecell{0.5919 \\(0.0160)}} & \multicolumn{1}{c}{{\makecell{-6.9530 \\(0.0122)}}} & {\textbf{\makecell{0.2153 \\(0.0076)}}} & \multicolumn{1}{c}{\makecell{0.0117 \\(0.0229)}} & \multicolumn{1}{c}{\makecell{0.0117 \\(0.0189)}} & \multicolumn{1}{c}{\makecell{0.0375 \\(0.0561)}} & \multicolumn{1}{c}{\makecell{0.0203 \\(0.0326)}} & \makecell{1.3120 \\(0.7623)}                 \\
\bottomrule
\end{tabular}
}
\label{tab:general_performance_SUPPORT_age_CI}
\vspace{-2em}
\end{table}

\begin{table}[H]
\caption{\small Test set scores on the SUPPORT (race) dataset, in the same format as Table~\ref{tab:general_performance_CI}.} 
\centering
\setlength\tabcolsep{3.6pt}
{\tiny %
\renewcommand{\belowrulesep}{0.1pt}
\renewcommand{\aboverulesep}{0.1pt}
\begin{tabular}{cccccc|ccccc}
\toprule
\multirow{2}{*}{} & \multirow{2}{*}{Methods} & \multicolumn{4}{c|}{Accuracy Metrics}                                                    & \multicolumn{5}{c}{Fairness Metrics}                             \\ \cmidrule{3-11}
&                   & \multicolumn{1}{c}{c-index$\uparrow$} & \multicolumn{1}{c}{AUC$\uparrow$} & \multicolumn{1}{c}{LPL$\uparrow$} & IBS$\downarrow$ & \multicolumn{1}{c}{F$_I$$\downarrow$} & \multicolumn{1}{c}{F$_G$$\downarrow$} & \multicolumn{1}{c}{F$_{\cap}$$\downarrow$}& \multicolumn{1}{c}{F$_A$$\downarrow$} &CI(\%)$\downarrow$                  \\ \midrule
\multirow{8}{*}{\rotatebox{90}{Linear \ \ \ \ \ \ \ \ \ \ \ \ \ \ }} &        \tableCox            & \multicolumn{1}{c}{{\textbf{\makecell{0.6025 \\(0.0005)}}}} & \multicolumn{1}{c}{\makecell{0.6163 \\(0.0010)}} & \multicolumn{1}{c}{{\textbf{\makecell{-6.8761 \\(0.0010)}}}} & {\makecell{0.2304 \\(0.0015)}} & \multicolumn{1}{c}{\makecell{0.2113 \\(0.0093)}} & \multicolumn{1}{c}{\makecell{0.0297 \\(0.0130)}} & \multicolumn{1}{c}{\makecell{0.4490 \\(0.0322)}} & \multicolumn{1}{c}{\makecell{0.2300 \\(0.0085)}} & \makecell{1.4160 \\(0.0696)}             \\  \cmidrule{3-11}
&          \tableCoxKeyaInd          & \multicolumn{1}{c}{{\makecell{0.5905 \\(0.0086)}}} & \multicolumn{1}{c}{\makecell{0.6007\\ (0.0108)}} & \multicolumn{1}{c}{{\makecell{-6.9390 \\(0.0214)}}} & {\textbf{\makecell{0.2161 \\(0.0054)}}} & \multicolumn{1}{c}{\makecell{0.0381\\ (0.0383)}} & \multicolumn{1}{c}{\makecell{0.0148 \\(0.0197)}} & \multicolumn{1}{c}{\makecell{0.0910 \\(0.0846)}} & \multicolumn{1}{c}{\makecell{0.0479 \\(0.0459)}} & \makecell{1.1230\\ (0.6621)}               \\
\bottomrule
\end{tabular}
}
\label{tab:general_performance_SUPPORT_race_CI}
\end{table}

\begin{table}[H]
\caption{\small Test set scores on the SEER (age) dataset, in the same format as Table~\ref{tab:general_performance_CI}.} 
\centering
\setlength\tabcolsep{3.6pt}
{\tiny %
\renewcommand{\belowrulesep}{0.1pt}
\renewcommand{\aboverulesep}{0.1pt}
\begin{tabular}{cccccc|ccccc}
\toprule
\multirow{2}{*}{} & \multirow{2}{*}{Methods} & \multicolumn{4}{c|}{Accuracy Metrics}                                                    & \multicolumn{5}{c}{Fairness Metrics}                             \\ \cmidrule{3-11}
&                   & \multicolumn{1}{c}{c-index$\uparrow$} & \multicolumn{1}{c}{AUC$\uparrow$} & \multicolumn{1}{c}{LPL$\uparrow$} & IBS$\downarrow$ & \multicolumn{1}{c}{F$_I$$\downarrow$} & \multicolumn{1}{c}{F$_G$$\downarrow$} & \multicolumn{1}{c}{F$_{\cap}$$\downarrow$}& \multicolumn{1}{c}{F$_A$$\downarrow$} &CI(\%)$\downarrow$                  \\ \midrule
\multirow{8}{*}{\rotatebox{90}{Linear \ \ \ \ \ \ \ \ \ \ \ \ \ \ }} &        \tableCox            & \multicolumn{1}{c}{{\makecell{0.7409 \\(0.0016)}}} & \multicolumn{1}{c}{\makecell{0.7624 \\(0.0017)}} & \multicolumn{1}{c}{{\textbf{\makecell{-5.9427 \\(0.0034)}}}} & {\makecell{0.0964 \\(0.0008)}} & \multicolumn{1}{c}{\makecell{0.6105 \\(0.0307)}} & \multicolumn{1}{c}{\makecell{0.9037 \\(0.1308)}} & \multicolumn{1}{c}{\makecell{0.6750 \\(0.0630)}} & \multicolumn{1}{c}{\makecell{0.7297 \\(0.0656)}} & \makecell{2.5640 \\(0.3531)}               \\  \cmidrule{3-11}
&          \tableCoxKeyaInd          & \multicolumn{1}{c}{{\makecell{0.7118 \\(0.0277)}}} & \multicolumn{1}{c}{\makecell{0.7345 \\(0.0287)}} & \multicolumn{1}{c}{{\makecell{-6.1539 \\(0.0716)}}} & {\makecell{0.0896 \\(0.0014)}} & \multicolumn{1}{c}{\makecell{0.2045 \\(0.0587)}} & \multicolumn{1}{c}{\makecell{0.2900 \\(0.1183)}} & \multicolumn{1}{c}{\makecell{0.3802 \\(0.1431)}} & \multicolumn{1}{c}{\makecell{0.2916 \\(0.0994)}} & \makecell{1.4310 \\(1.2077)}                \\
\bottomrule
\end{tabular}
}
\label{tab:general_performance_SEER_age_CI}
\end{table}

\begin{table}[H]
\caption{\small Test set scores on the FLC (age) dataset when hyperparameter tuning is based on F$_A$. The format of this table is the same that of Table~\ref{tab:general_performance_CI}.}\vspace{-1em}
\centering
\setlength\tabcolsep{3.6pt}
{\tiny %
\renewcommand{\belowrulesep}{0.1pt}
\renewcommand{\aboverulesep}{0.1pt}
\begin{tabular}{cccccc|ccccc}
\toprule
\multirow{2}{*}{} & \multirow{2}{*}{Methods} & \multicolumn{4}{c|}{Accuracy Metrics}                                                    & \multicolumn{5}{c}{Fairness Metrics}                             \\ \cmidrule{3-11}
                   &                   & \multicolumn{1}{c}{c-index$\uparrow$} & \multicolumn{1}{c}{AUC$\uparrow$} & \multicolumn{1}{c}{LPL$\uparrow$} & IBS$\downarrow$ & \multicolumn{1}{c}{F$_I$$\downarrow$} & \multicolumn{1}{c}{F$_G$$\downarrow$} & \multicolumn{1}{c}{F$_{\cap}$$\downarrow$}& \multicolumn{1}{c}{F$_A$$\downarrow$} &CI(\%)$\downarrow$                  \\ \midrule
\multirow{8}{*}{\rotatebox{90}{Linear \ \ \ \ \ \ \ \ \ \ \ \ \ \ }} &        \tableCox            & \multicolumn{1}{c}{\textbf{\makecell{0.8032 \\(0.0002)}}} & \multicolumn{1}{c}{\makecell{0.8176 \\(0.0005)}} & \multicolumn{1}{c}{\textbf{\makecell{-6.3724 \\(0.0011)}}} & {\makecell{0.1739 \\(0.0004)}} & \multicolumn{1}{c}{\makecell{1.8787 \\(0.0304)}} & \multicolumn{1}{c}{\makecell{3.0282 \\(0.0469)}} & \multicolumn{1}{c}{\makecell{2.8355 \\(0.0297)}} & \multicolumn{1}{c}{\makecell{2.5808 \\(0.0332)}} & \makecell{0.5350 \\(0.0413)}              \\ \cmidrule{3-11}
&          \tableCoxKeyaInd          & \multicolumn{1}{c}{{\makecell{0.7852 \\(0.0220)}}} & \multicolumn{1}{c}{\makecell{0.8071 \\(0.0240)}} & \multicolumn{1}{c}{\makecell{-6.7714 \\(0.0360)}} & {\makecell{0.1333 \\(0.0034)}} & \multicolumn{1}{c}{\makecell{0.2233 \\(0.0161)}} & \multicolumn{1}{c}{\makecell{0.3809 \\(0.0317)}} & \multicolumn{1}{c}{\makecell{0.7013 \\(0.0406)}} & \multicolumn{1}{c}{\makecell{0.4352 \\(0.0222)}} & \makecell{0.6500 \\(0.6830)}                 \\
\bottomrule
\end{tabular}
}
\label{tab:general_performance}
\vspace{-2em}
\end{table}

\begin{table}[H]
\caption{\small Test set scores on the FLC (gender) dataset when hyperparameter tuning is based on F$_A$. The format of this table is the same that of Table~\ref{tab:general_performance_CI}.}\vspace{-1em} 
\centering
\setlength\tabcolsep{3.6pt}
{\tiny %
\renewcommand{\belowrulesep}{0.1pt}
\renewcommand{\aboverulesep}{0.1pt}
\begin{tabular}{cccccc|ccccc}
\toprule
\multirow{2}{*}{} & \multirow{2}{*}{Methods} & \multicolumn{4}{c|}{Accuracy Metrics}                                                    & \multicolumn{5}{c}{Fairness Metrics}                             \\ \cmidrule{3-11}
&                   & \multicolumn{1}{c}{c-index$\uparrow$} & \multicolumn{1}{c}{AUC$\uparrow$} & \multicolumn{1}{c}{LPL$\uparrow$} & IBS$\downarrow$ & \multicolumn{1}{c}{F$_I$$\downarrow$} & \multicolumn{1}{c}{F$_G$$\downarrow$} & \multicolumn{1}{c}{F$_{\cap}$$\downarrow$}& \multicolumn{1}{c}{F$_A$$\downarrow$} &CI(\%)$\downarrow$                  \\ \midrule
\multirow{8}{*}{\rotatebox{90}{Linear \ \ \ \ \ \ \ \ \ \ \ \ \ \ }} &        \tableCox            & \multicolumn{1}{c}{{{\textbf{\makecell{0.8032 \\(0.0002)}}}}} & \multicolumn{1}{c}{{\makecell{0.8176 \\(0.0005)}}} & \multicolumn{1}{c}{{\textbf{\makecell{-6.3724 \\(0.0011)}}}} & {\makecell{0.1739 \\(0.0004)}} & \multicolumn{1}{c}{\makecell{1.8787 \\(0.0304)}} & \multicolumn{1}{c}{\makecell{0.5421 \\(0.0299)}} & \multicolumn{1}{c}{\makecell{2.8355 \\(0.0297)}} & \multicolumn{1}{c}{\makecell{1.7521 \\(0.0266)}} & \makecell{0.8610 \\(0.0197)}               \\  \cmidrule{3-11}
&          \tableCoxKeyaInd          & \multicolumn{1}{c}{{{\makecell{0.7918 \\(0.0078)}}}} & \multicolumn{1}{c}{{\makecell{0.8158 \\(0.0068)}}} & \multicolumn{1}{c}{{\makecell{-6.7585 \\(0.0202)}}} & {\makecell{0.1335 \\(0.0035)}} & \multicolumn{1}{c}{\makecell{0.2282 \\(0.0164)}} & \multicolumn{1}{c}{\makecell{0.0190 \\(0.0111)}} & \multicolumn{1}{c}{\makecell{0.6974 \\(0.0372)}} & \multicolumn{1}{c}{\makecell{0.3149 \\(0.0194)}} & \makecell{1.6720 \\(0.8430)}                  \\
\bottomrule
\end{tabular}
}
\label{tab:general_performance_FLC_gender}
\end{table}

\begin{table}[H]
\caption{\small Test set scores on the SUPPORT (age) dataset when hyperparameter tuning is based on F$_A$. The format of this table is the same that of Table~\ref{tab:general_performance_CI}.}\vspace{-1em}
\centering
\setlength\tabcolsep{3.6pt}
{\tiny %
\renewcommand{\belowrulesep}{0.1pt}
\renewcommand{\aboverulesep}{0.1pt}
\begin{tabular}{cccccc|ccccc}
\toprule
\multirow{2}{*}{} & \multirow{2}{*}{Methods} & \multicolumn{4}{c|}{Accuracy Metrics}                                                    & \multicolumn{5}{c}{Fairness Metrics}                             \\ \cmidrule{3-11}
&                   & \multicolumn{1}{c}{c-index$\uparrow$} & \multicolumn{1}{c}{AUC$\uparrow$} & \multicolumn{1}{c}{LPL$\uparrow$} & IBS$\downarrow$ & \multicolumn{1}{c}{F$_I$$\downarrow$} & \multicolumn{1}{c}{F$_G$$\downarrow$} & \multicolumn{1}{c}{F$_{\cap}$$\downarrow$}& \multicolumn{1}{c}{F$_A$$\downarrow$} &CI(\%)$\downarrow$                  \\ \midrule
\multirow{8}{*}{\rotatebox{90}{Linear \ \ \ \ \ \ \ \ \ \ \ \ \ \ }} &        \tableCox            & \multicolumn{1}{c}{\textbf{\makecell{0.6025 \\(0.0005)}}} & \multicolumn{1}{c}{\makecell{0.6163 \\(0.0010)}} & \multicolumn{1}{c}{\textbf{\makecell{-6.8761 \\(0.0010)}}} & \textbf{\makecell{0.2304 \\(0.0015)}} & \multicolumn{1}{c}{\makecell{0.2113 \\(0.0093)}} & \multicolumn{1}{c}{\makecell{0.1528 \\(0.0059)}} & \multicolumn{1}{c}{\makecell{0.4490 \\(0.0322)}} & \multicolumn{1}{c}{\makecell{0.2710 \\(0.0128)}} & \makecell{2.2240 \\(0.1078)}              \\  \cmidrule{3-11}
&          \tableCoxKeyaInd          & \multicolumn{1}{c}{{\makecell{0.5831 \\(0.0100)}}} & \multicolumn{1}{c}{\makecell{0.5932 \\(0.0143)}} & \multicolumn{1}{c}{\makecell{-6.9594 \\(0.0054)}} & {\textbf{\makecell{0.2147 \\(0.0063)}}} & \multicolumn{1}{c}{\makecell{0.0011 \\(0.0001)}} & \multicolumn{1}{c}{\makecell{0.0038 \\(0.0010)}} & \multicolumn{1}{c}{\makecell{0.0180 \\(0.0036)}} & \multicolumn{1}{c}{\makecell{0.0076 \\(0.0015)}} & \makecell{1.4280 \\(0.8245)}                 \\
\bottomrule
\end{tabular}
}
\label{tab:general_performance_SUPPORT_age}
\vspace{-2em}
\end{table}

\begin{table}[H]
\caption{\small Test set scores on the SUPPORT (race) dataset when hyperparameter tuning is based on F$_A$. The format of this table is the same that of Table~\ref{tab:general_performance_CI}.}\vspace{-1em}
\centering
\setlength\tabcolsep{3.6pt}
{\tiny %
\renewcommand{\belowrulesep}{0.1pt}
\renewcommand{\aboverulesep}{0.1pt}
\begin{tabular}{cccccc|ccccc}
\toprule
\multirow{2}{*}{} & \multirow{2}{*}{Methods} & \multicolumn{4}{c|}{Accuracy Metrics}                                                    & \multicolumn{5}{c}{Fairness Metrics}                             \\ \cmidrule{3-11}
&                   & \multicolumn{1}{c}{c-index$\uparrow$} & \multicolumn{1}{c}{AUC$\uparrow$} & \multicolumn{1}{c}{LPL$\uparrow$} & IBS$\downarrow$ & \multicolumn{1}{c}{F$_I$$\downarrow$} & \multicolumn{1}{c}{F$_G$$\downarrow$} & \multicolumn{1}{c}{F$_{\cap}$$\downarrow$}& \multicolumn{1}{c}{F$_A$$\downarrow$} &CI(\%)$\downarrow$                  \\ \midrule
\multirow{8}{*}{\rotatebox{90}{Linear \ \ \ \ \ \ \ \ \ \ \ \ \ \ }} &        \tableCox            & \multicolumn{1}{c}{{{\makecell{0.6025 \\(0.0005)}}}} & \multicolumn{1}{c}{{\makecell{0.6163 \\(0.0010)}}} & \multicolumn{1}{c}{{\textbf{\makecell{-6.8761 \\(0.0010)}}}} & {\makecell{0.2304 \\(0.0015)}} & \multicolumn{1}{c}{\makecell{0.2113 \\(0.0093)}} & \multicolumn{1}{c}{\makecell{0.0297 \\(0.0130)}} & \multicolumn{1}{c}{\makecell{0.4490 \\(0.0322)}} & \multicolumn{1}{c}{\makecell{0.2300 \\(0.0085)}} & \makecell{1.4160 \\(0.0696)}               \\  \cmidrule{3-11}
&          \tableCoxKeyaInd          & \multicolumn{1}{c}{{{\makecell{0.5831 \\(0.0100)}}}} & \multicolumn{1}{c}{{\makecell{0.5932 \\(0.0143)}}} & \multicolumn{1}{c}{{\makecell{-6.9594 \\(0.0054)}}} & {\textbf{\makecell{0.2147 \\(0.0063)}}} & \multicolumn{1}{c}{\textbf{\makecell{0.0001 \\(0.0001)}}} & \multicolumn{1}{c}{\makecell{0.0023 \\(0.0016)}} & \multicolumn{1}{c}{\makecell{0.0080 \\(0.0036)}} & \multicolumn{1}{c}{\makecell{0.0034 \\(0.0017)}} & \makecell{1.2580 \\(0.5410)}                  \\
\bottomrule
\end{tabular}
}
\label{tab:general_performance_SUPPORT_race}
\end{table}

\begin{table}[H]
\caption{\small Test set scores on the SUPPORT (gender) dataset when hyperparameter tuning is based on F$_A$. The format of this table is the same that of Table~\ref{tab:general_performance_CI}.}\vspace{-1em}
\centering
\setlength\tabcolsep{3.6pt}
{\tiny %
\renewcommand{\belowrulesep}{0.1pt}
\renewcommand{\aboverulesep}{0.1pt}
\begin{tabular}{cccccc|ccccc}
\toprule
\multirow{2}{*}{} & \multirow{2}{*}{Methods} & \multicolumn{4}{c|}{Accuracy Metrics}                                                    & \multicolumn{5}{c}{Fairness Metrics}                             \\ \cmidrule{3-11}
&                   & \multicolumn{1}{c}{c-index$\uparrow$} & \multicolumn{1}{c}{AUC$\uparrow$} & \multicolumn{1}{c}{LPL$\uparrow$} & IBS$\downarrow$ & \multicolumn{1}{c}{F$_I$$\downarrow$} & \multicolumn{1}{c}{F$_G$$\downarrow$} & \multicolumn{1}{c}{F$_{\cap}$$\downarrow$}& \multicolumn{1}{c}{F$_A$$\downarrow$} &CI(\%)$\downarrow$                  \\ \midrule
\multirow{8}{*}{\rotatebox{90}{Linear \ \ \ \ \ \ \ \ \ \ \ \ \ \ }} &        \tableCox            & \multicolumn{1}{c}{{{\makecell{0.6025 \\(0.0005)}}}} & \multicolumn{1}{c}{{\makecell{0.6163 \\(0.0010)}}} & \multicolumn{1}{c}{{\textbf{\makecell{-6.8761 \\(0.0010)}}}} & {\makecell{0.2304 \\(0.0015)}} & \multicolumn{1}{c}{\makecell{0.2113 \\(0.0093)}} & \multicolumn{1}{c}{\makecell{0.0439 \\(0.0052)}} & \multicolumn{1}{c}{\makecell{0.4490 \\(0.0322)}} & \multicolumn{1}{c}{\makecell{0.2347 \\(0.0127)}} & \makecell{1.4300 \\(0.0654)}               \\  \cmidrule{3-11}
&          \tableCoxKeyaInd          & \multicolumn{1}{c}{{{\makecell{0.5831 \\(0.0100)}}}} & \multicolumn{1}{c}{{\makecell{0.5932 \\(0.0143)}}} & \multicolumn{1}{c}{{\makecell{-6.9594 \\(0.0054)}}} & {\textbf{\makecell{0.2147 \\(0.0063)}}} & \multicolumn{1}{c}{\textbf{\makecell{0.0001 \\(0.0001)}}} & \multicolumn{1}{c}{\makecell{0.0012 \\(0.0004)}} & \multicolumn{1}{c}{\makecell{0.0081 \\(0.0036)}} & \multicolumn{1}{c}{\makecell{0.0031 \\(0.0013)}} & \makecell{1.1110 \\(0.7139)}                  \\
\bottomrule
\end{tabular}
}
\label{tab:general_performance_SUPPORT_gender}
\end{table}

\begin{table}[H]
\caption{\small Test set scores on the SEER (age) dataset when hyperparameter tuning is based on F$_A$. The format of this table is the same that of Table~\ref{tab:general_performance_CI}.}\vspace{-1em}
\centering
\setlength\tabcolsep{3.6pt}
{\tiny %
\renewcommand{\belowrulesep}{0.1pt}
\renewcommand{\aboverulesep}{0.1pt}
\begin{tabular}{cccccc|ccccc}
\toprule
\multirow{2}{*}{} & \multirow{2}{*}{Methods} & \multicolumn{4}{c|}{Accuracy Metrics}                                                    & \multicolumn{5}{c}{Fairness Metrics}                             \\ \cmidrule{3-11}
&                   & \multicolumn{1}{c}{c-index$\uparrow$} & \multicolumn{1}{c}{AUC$\uparrow$} & \multicolumn{1}{c}{LPL$\uparrow$} & IBS$\downarrow$ & \multicolumn{1}{c}{F$_I$$\downarrow$} & \multicolumn{1}{c}{F$_G$$\downarrow$} & \multicolumn{1}{c}{F$_{\cap}$$\downarrow$}& \multicolumn{1}{c}{F$_A$$\downarrow$} &CI(\%)$\downarrow$                  \\ \midrule
\multirow{8}{*}{\rotatebox{90}{Linear \ \ \ \ \ \ \ \ \ \ \ \ \ \ }} &        \tableCox            & \multicolumn{1}{c}{{\textbf{\makecell{0.7409 \\(0.0016)}}}} & \multicolumn{1}{c}{\textbf{\makecell{0.7624 \\(0.0017)}}} & \multicolumn{1}{c}{\textbf{\makecell{-5.9427 \\(0.0034)}}} & {\makecell{0.0964 \\(0.0008)}} & \multicolumn{1}{c}{\makecell{0.6105 \\(0.0307)}} & \multicolumn{1}{c}{\makecell{0.9037 \\(0.1308)}} & \multicolumn{1}{c}{\makecell{0.6750 \\(0.0630)}} & \multicolumn{1}{c}{\makecell{0.7297 \\(0.0656)}} & \makecell{2.5640 \\(0.3531)}               \\  \cmidrule{3-11}
&          \tableCoxKeyaInd          & \multicolumn{1}{c}{{\makecell{0.7229 \\(0.0177)}}} & \multicolumn{1}{c}{\makecell{0.7473 \\(0.0207)}} & \multicolumn{1}{c}{\makecell{-6.1913 \\(0.0250)}} & {\makecell{0.0885 \\(0.0009)}} & \multicolumn{1}{c}{\makecell{0.1299 \\(0.0177)}} & \multicolumn{1}{c}{\makecell{0.1312 \\(0.0546)}} & \multicolumn{1}{c}{\makecell{0.1824 \\(0.0765)}} & \multicolumn{1}{c}{\makecell{0.1479\\ (0.0432)}} & \makecell{1.4070 \\(1.0039)}                 \\
\bottomrule
\end{tabular}
}
\label{tab:general_performance_SEER_age}
\end{table}

\begin{table}[H]
\caption{\small Test set scores on the SEER (race) dataset when hyperparameter tuning is based on F$_A$. The format of this table is the same that of Table~\ref{tab:general_performance_CI}.}\vspace{-1em}
\centering
\setlength\tabcolsep{3.6pt}
{\tiny %
\renewcommand{\belowrulesep}{0.1pt}
\renewcommand{\aboverulesep}{0.1pt}
\begin{tabular}{cccccc|ccccc}
\toprule
\multirow{2}{*}{} & \multirow{2}{*}{Methods} & \multicolumn{4}{c|}{Accuracy Metrics}                                                    & \multicolumn{5}{c}{Fairness Metrics}                             \\ \cmidrule{3-11}
&                   & \multicolumn{1}{c}{c-index$\uparrow$} & \multicolumn{1}{c}{AUC$\uparrow$} & \multicolumn{1}{c}{LPL$\uparrow$} & IBS$\downarrow$ & \multicolumn{1}{c}{F$_I$$\downarrow$} & \multicolumn{1}{c}{F$_G$$\downarrow$} & \multicolumn{1}{c}{F$_{\cap}$$\downarrow$}& \multicolumn{1}{c}{F$_A$$\downarrow$} &CI(\%)$\downarrow$                  \\ \midrule
\multirow{8}{*}{\rotatebox{90}{Linear \ \ \ \ \ \ \ \ \ \ \ \ \ \ }} &        \tableCox            & \multicolumn{1}{c}{{{\textbf{\makecell{0.7409 \\(0.0016)}}}}} & \multicolumn{1}{c}{{\textbf{\makecell{0.7624 \\(0.0017)}}}} & \multicolumn{1}{c}{{\textbf{\makecell{-5.9427 \\(0.0034)}}}} & {\makecell{0.0964 \\(0.0008)}} & \multicolumn{1}{c}{\makecell{0.6105 \\(0.0307)}} & \multicolumn{1}{c}{\makecell{0.1183\\ (0.0645)}} & \multicolumn{1}{c}{\makecell{0.6750\\ (0.0630)}} & \multicolumn{1}{c}{\makecell{0.4679 \\(0.0437)}} & \makecell{1.1880 \\(0.1742)}               \\  \cmidrule{3-11}
&          \tableCoxKeyaInd          & \multicolumn{1}{c}{{{\makecell{0.7224 \\(0.0176)}}}} & \multicolumn{1}{c}{{\makecell{0.7471 \\(0.0208)}}} & \multicolumn{1}{c}{{\makecell{-6.1914 \\(0.0252)}}} & {\makecell{0.0885 \\(0.0009)}} & \multicolumn{1}{c}{\makecell{0.1292 \\(0.0181)}} & \multicolumn{1}{c}{\makecell{0.0318\\ (0.0280)}} & \multicolumn{1}{c}{\makecell{0.1828 \\(0.0760)}} & \multicolumn{1}{c}{\makecell{0.1146 \\(0.0309)}} & \makecell{2.1470 \\(1.6144)}                  \\
\bottomrule
\end{tabular}
}
\label{tab:general_performance_SEER_race}
\end{table}

\begin{table}[H]
\caption{\small Test set scores for \textsc{dro-cox (split)} on the FLC (age) dataset using $n_2=0.1n, 0.2n, 0.3n, 0.4n, 0.5n$ (corresponding to $n_1=0.9n,0.8n,0.7n,0.6n,0.5n$). The format of this table is similar to that of Table~\ref{tab:general_performance_CI} although here we do not bold or highlight any cells, as our main finding here is that the scores are not dramatically different for the different choices for $n_1$ or $n_2$.}
\vspace{-1em}
\centering
\setlength\tabcolsep{8pt}
{\tiny %
\renewcommand{\belowrulesep}{0.1pt}
\renewcommand{\aboverulesep}{0.1pt}
\begin{tabular}{cccccc|ccccc}
\toprule
\multirow{2}{*}{} & \multirow{2}{*}{$n_2$} & \multicolumn{4}{c|}{Accuracy Metrics}                                                    & \multicolumn{5}{c}{Fairness Metrics}                             \\ \cmidrule{3-11}
&                   & \multicolumn{1}{c}{c-index$\uparrow$} & \multicolumn{1}{c}{AUC$\uparrow$} & \multicolumn{1}{c}{LPL$\uparrow$} & IBS$\downarrow$ & \multicolumn{1}{c}{F$_I$$\downarrow$} & \multicolumn{1}{c}{F$_G$$\downarrow$} & \multicolumn{1}{c}{F$_{\cap}$$\downarrow$}& \multicolumn{1}{c}{F$_A$$\downarrow$} &CI(\%)$\downarrow$                  \\ \midrule
\multirow{5}{*}{\rotatebox{90}{Linear \ \ \ \ \ \ \ \ }} &        $0.1n$           & \multicolumn{1}{c}{{\makecell{0.7813 \\(0.0181)}}} & \multicolumn{1}{c}{\makecell{0.8023 \\(0.0174)}} & \multicolumn{1}{c}{{\makecell{-7.0822 \\(0.0069)}}} & {\makecell{0.1415 \\(0.0063)}} & \multicolumn{1}{c}{\makecell{0.0035 \\(0.0030)}} & \multicolumn{1}{c}{\makecell{0.0118 \\(0.0062)}} & \multicolumn{1}{c}{\makecell{0.0292 \\(0.0138)}} & \multicolumn{1}{c}{\makecell{0.0148 \\(0.0075)}} & \makecell{0.5670 \\(0.3535)}              \\ \cmidrule{3-11}
&          $0.2n$         & \multicolumn{1}{c}{{\makecell{0.7955 \\(0.0053)}}} & \multicolumn{1}{c}{\makecell{0.8156 \\(0.0036)}} & \multicolumn{1}{c}{{\makecell{-7.0835 \\(0.0044)}}} & {\makecell{0.1403 \\(0.0031)}} & \multicolumn{1}{c}{\makecell{0.0026 \\(0.0020)}} & \multicolumn{1}{c}{\makecell{0.0105 \\(0.0032)}} & \multicolumn{1}{c}{\makecell{0.0251 \\(0.0107)}} & \multicolumn{1}{c}{\makecell{0.0127 \\(0.0053)}} & \makecell{0.3810 \\(0.2621)}                 \\ \cmidrule{3-11}
&          $0.3n$         & \multicolumn{1}{c}{{\makecell{0.7976 \\(0.0027)}}} & \multicolumn{1}{c}{\makecell{0.8181 \\(0.0026)}} & \multicolumn{1}{c}{{\makecell{-7.0844 \\(0.0033)}}} & {\makecell{0.1398 \\(0.0025)}} & \multicolumn{1}{c}{\makecell{0.0021 \\(0.0015)}} & \multicolumn{1}{c}{\makecell{0.0099 \\(0.0025)}} & \multicolumn{1}{c}{\makecell{0.0254 \\(0.0074)}} & \multicolumn{1}{c}{\makecell{0.0124 \\(0.0038)}} & \makecell{0.2150 \\(0.1907)}                 \\ \cmidrule{3-11}
&       $0.4n$           & \multicolumn{1}{c}{{\makecell{0.7969 \\(0.0040)}}} & \multicolumn{1}{c}{\makecell{0.8174 \\(0.0024)}} & \multicolumn{1}{c}{{\makecell{-7.0852 \\(0.0019)}}} & {\makecell{0.1393 \\(0.0014)}} & \multicolumn{1}{c}{\makecell{0.0018 \\(0.0008)}} & \multicolumn{1}{c}{\makecell{0.0093 \\(0.0014)}} & \multicolumn{1}{c}{\makecell{0.0240 \\(0.0043)}} & \multicolumn{1}{c}{\makecell{0.0117 \\(0.0021)}} & \makecell{0.2910 \\(0.1280)}                \\ \cmidrule{3-11}
&       $0.5n$           & \multicolumn{1}{c}{{\makecell{0.7963 \\(0.0045)}}} & \multicolumn{1}{c}{\makecell{0.8168 \\(0.0030)}} & \multicolumn{1}{c}{\makecell{-7.0856 \\(0.0009)}} & {\makecell{0.1390 \\(0.0008)}} & \multicolumn{1}{c}{\makecell{0.0016 \\(0.0004)}} & \multicolumn{1}{c}{\makecell{0.0089 \\(0.0008)}} & \multicolumn{1}{c}{\makecell{0.0232 \\(0.0031)}} & \multicolumn{1}{c}{\makecell{0.0113 \\(0.0013)}} & \makecell{0.2340 \\(0.1237)}                \\ \cmidrule{1-11}
\multirow{5}{*}{\rotatebox{90}{Nonlinear \ \ \ \ \ \   }} &        $0.1n$           & \multicolumn{1}{c}{{\makecell{0.7619 \\(0.0068)}}} & \multicolumn{1}{c}{\makecell{0.7707 \\(0.0079)}} & \multicolumn{1}{c}{{\makecell{-6.8103 \\(0.0186)}}} & {\makecell{0.1703 \\(0.0002)}} & \multicolumn{1}{c}{\makecell{0.3923 \\(0.0514)}} & \multicolumn{1}{c}{\makecell{0.4897 \\(0.0509)}} & \multicolumn{1}{c}{\makecell{0.6953 \\(0.0889)}} & \multicolumn{1}{c}{\makecell{0.5258 \\(0.0624)}} & \makecell{2.8090 \\(0.1807)}                \\  \cmidrule{3-11}
&          $0.2n$         & \multicolumn{1}{c}{{\makecell{0.7621 \\(0.0069)}}} & \multicolumn{1}{c}{\makecell{0.7710 \\(0.0082)}} & \multicolumn{1}{c}{{\makecell{-6.8114 \\(0.0186)}}} & {\makecell{0.1703 \\(0.0002)}} & \multicolumn{1}{c}{\makecell{0.4167 \\(0.0676)}} & \multicolumn{1}{c}{\makecell{0.5056 \\(0.0591)}} & \multicolumn{1}{c}{\makecell{0.7299 \\(0.1075)}} & \multicolumn{1}{c}{\makecell{0.5507 \\(0.0773)}} & \makecell{2.8030 \\(0.2261)}                 \\ \cmidrule{3-11}
&          $0.3n$         & \multicolumn{1}{c}{{\makecell{0.7627 \\(0.0067)}}} & \multicolumn{1}{c}{\makecell{0.7719 \\(0.0080)}} & \multicolumn{1}{c}{{\makecell{-6.8115 \\(0.0182)}}} & {\makecell{0.1703 \\(0.0002)}} & \multicolumn{1}{c}{\makecell{0.4321 \\(0.0755)}} & \multicolumn{1}{c}{\makecell{0.5164 \\(0.0645)}} & \multicolumn{1}{c}{\makecell{0.7525 \\(0.1159)}} & \multicolumn{1}{c}{\makecell{0.5670 \\(0.0849)}} & \makecell{2.7770 \\(0.2414)}                 \\ \cmidrule{3-11}
&       $0.4n$           & \multicolumn{1}{c}{{\makecell{0.7627 \\(0.0061)}}} & \multicolumn{1}{c}{\makecell{0.7719 \\(0.0073)}} & \multicolumn{1}{c}{{\makecell{-6.8123 \\(0.0180)}}} & {\makecell{0.1703 \\(0.0002)}} & \multicolumn{1}{c}{\makecell{0.4414 \\(0.1018)}} & \multicolumn{1}{c}{\makecell{0.5236 \\(0.0798)}} & \multicolumn{1}{c}{\makecell{0.7578 \\(0.1316)}} & \multicolumn{1}{c}{\makecell{0.5742 \\(0.1037)}} & \makecell{2.7930 \\(0.2281)}                \\ \cmidrule{3-11}
&       $0.5n$         & \multicolumn{1}{c}{{\makecell{0.7629 \\(0.0064)}}} & \multicolumn{1}{c}{\makecell{0.7719 \\(0.0076)}} & \multicolumn{1}{c}{\makecell{-6.8131 \\(0.0199)}} & {\makecell{0.1703 \\(0.0002)}} & \multicolumn{1}{c}{\makecell{0.4347 \\(0.1214)}} & \multicolumn{1}{c}{\makecell{0.5184 \\(0.0922)}} & \multicolumn{1}{c}{\makecell{0.7508 \\(0.1417)}} & \multicolumn{1}{c}{\makecell{0.5680 \\(0.1176)}} & \makecell{2.8490 \\(0.2435)}                 \\ \bottomrule
\end{tabular}
}
\label{tab:performance_of_diff_split}
\vspace{-1em}
\end{table}

\begin{table}[H]
\caption{\small Test set scores of \textsc{dro-cox (split, one side)} vs \textsc{dro-cox (split)} on the FLC (age) dataset. The format of this table is the same that of Table~\ref{tab:general_performance_CI} except without any cells highlighted in green as we are not comparing against baselines by previous authors.}\vspace{-1em}
\centering
\setlength\tabcolsep{2pt}
{\tiny %
\renewcommand{\belowrulesep}{0.1pt}
\renewcommand{\aboverulesep}{0.1pt}
\begin{tabular}{cccccc|ccccc}
\toprule
\multirow{2}{*}{} & \multirow{2}{*}{Methods} & \multicolumn{4}{c|}{Accuracy Metrics}                                                    & \multicolumn{5}{c}{Fairness Metrics}                             \\ \cmidrule{3-11}
&                   & \multicolumn{1}{c}{c-index$\uparrow$} & \multicolumn{1}{c}{AUC$\uparrow$} & \multicolumn{1}{c}{LPL$\uparrow$} & IBS$\downarrow$ & \multicolumn{1}{c}{F$_I$$\downarrow$} & \multicolumn{1}{c}{F$_G$$\downarrow$} & \multicolumn{1}{c}{F$_{\cap}$$\downarrow$}& \multicolumn{1}{c}{F$_A$$\downarrow$} &CI(\%)$\downarrow$                  \\ \midrule
\multirow{2}{*}{\rotatebox{90}{Linear }} &        \textsc{DRO-COX (SPLIT, ONE SIDE)}           & \multicolumn{1}{c}{{\makecell{0.7809 \\(0.0092)}}} & \multicolumn{1}{c}{\makecell{0.8031 \\(0.0101)}} & \multicolumn{1}{c}{{\makecell{-7.0862 \\(0.0029)}}} & {\textbf{\makecell{0.1330 \\(0.0002)}}} & \multicolumn{1}{c}{\makecell{0.0017 \\(0.0011)}} & \multicolumn{1}{c}{{\makecell{0.0090 \\(0.0024)}}} & \multicolumn{1}{c}{\makecell{0.0232 \\(0.0079)}} & \multicolumn{1}{c}{{\makecell{0.0113 \\(0.0037)}}} & \makecell{0.4420 \\(0.3206)}              \\ \cmidrule{3-11}
&       \tableDROCoxSplit            & \multicolumn{1}{c}{{\textbf{\makecell{0.7963 \\(0.0045)}}}} & \multicolumn{1}{c}{\textbf{\makecell{0.8168 \\(0.0030)}}} & \multicolumn{1}{c}{\textbf{\makecell{-7.0856 \\(0.0009)}}} & {\makecell{0.1390 \\(0.0008)}} & \multicolumn{1}{c}{\textbf{\makecell{0.0016 \\(0.0004)}}} & \multicolumn{1}{c}{\textbf{\makecell{0.0089 \\(0.0008)}}} & \multicolumn{1}{c}{\textbf{\makecell{0.0232 \\(0.0031)}}} & \multicolumn{1}{c}{\textbf{\makecell{0.0113 \\(0.0013)}}} & \textbf{\makecell{0.2340 \\(0.1237)}}                \\ \cmidrule{1-11}
\multirow{2}{*}{\rotatebox{90}{\makecell{Non-\\linear}}} &        Deep \textsc{DRO-COX (SPLIT, ONE SIDE)}           & \multicolumn{1}{c}{{\makecell{0.7625 \\(0.0062)}}} & \multicolumn{1}{c}{\makecell{0.7715 \\(0.0075)}} & \multicolumn{1}{c}{{\makecell{-6.8133 \\(0.0208)}}} & {\textbf{\makecell{0.1371 \\(0.0008)}}} & \multicolumn{1}{c}{\makecell{0.4402 \\(0.1369)}} & \multicolumn{1}{c}{\makecell{0.5232 \\(0.1039)}} & \multicolumn{1}{c}{{\makecell{0.7533 \\(0.1514)}}} & \multicolumn{1}{c}{\makecell{0.5722 \\(0.1298)}} & \textbf{\makecell{2.8000 \\(0.2671)}}                \\  \cmidrule{3-11}
&       \tableDeepDROCoxSplit          & \multicolumn{1}{c}{{\textbf{\makecell{0.7629 \\(0.0064)}}}} & \multicolumn{1}{c}{\textbf{\makecell{0.7719 \\(0.0076)}}} & \multicolumn{1}{c}{\textbf{\makecell{-6.8131 \\(0.0199)}}} & {\makecell{0.1703 \\(0.0002)}} & \multicolumn{1}{c}{\textbf{\makecell{0.4347 \\(0.1214)}}} & \multicolumn{1}{c}{\textbf{\makecell{0.5184 \\(0.0922)}}} & \multicolumn{1}{c}{\textbf{\makecell{0.7508 \\(0.1417)}}} & \multicolumn{1}{c}{\textbf{\makecell{0.5680 \\(0.1176)}}} & \makecell{2.8490 \\(0.2435)}                 \\ \bottomrule
\end{tabular}
}
\label{tab:performance_of_one_side_split}
\end{table}

\end{document}